\documentclass{article}

\usepackage{arxiv}

\usepackage[utf8]{inputenc} 
\usepackage[T1]{fontenc}    
\usepackage{hyperref}       
\usepackage{url}            
\usepackage{booktabs}       
\usepackage{amsfonts}       
\usepackage{microtype}      
\usepackage{lipsum}
\usepackage{graphicx}
\usepackage{tabularx}
\graphicspath{ {./figures/} }

\usepackage{subcaption} 
\usepackage{xspace}
\usepackage{amsmath}
\usepackage{siunitx}

\title{
   Does Demand Response Increase Vulnerability to Cyber Attacks by Adversarial Data Modifications?
}

\author{
  Clemens Kortmann \\
  Process Systems Engineering (AVT.SVT)\\
  RWTH Aachen University\\
  \texttt{clemens.kortmann@rwth-aachen.de} \\
   \And
   Eike Cramer \\
  Department of Chemical Engineering\\
  Sargent Centre for Process Systems Engineering\\
  University College London\\
  \texttt{eike.cramer@alumni.tu-berlin.de} \\
  \textit{Corresponding author}
}

\begin{document}
	
\maketitle

\begin{abstract}
Adversarial attacks are crafted data manipulations that aim to deteriorate the outcomes of prediction or decision-making algorithms.
In the energy systems literature, adversarial attacks have been studied with a focus on problems regarding the electricity grid. 
Such problems include forecasting and grid state estimation, where adversarial attacks are also known as false data injection attacks.
Only few studies have analyzed the potential impact that adversarial attacks have on the demand side. 
We analyze how manipulated price forecasts impact the decision-making in industrial demand response. 
To this end, we design adversarial attacks that aim to deteriorate the output of electricity price forecasting models and solve scheduling optimization problems of energy-intensive production processes using the distorted price forecasts. 
We make use of a generalized process model to investigate the vulnerability to adversarial attacks for a range of production scheduling problems with different levels of process flexibility.
We find that adversarial attacks can erode the profits gained from demand response. 
However, when perturbations are limited in extent (so that they are hard to detect by the human user), demand response preserves about 90\% of its financial advantage compared to steady-state process operation. 
Further, we find that the impact of adversarial attacks on demand response does not only depend on the magnitude of the perturbations but rather on the orientation of the adversarial perturbations. 
Therefore, we argue that attack analyses should explicitly incorporate the sensitivities of scheduling optimization models into the attack design to enable more rigorous assessments of decision-making under adversarial attacks.
\end{abstract}

\section{Introduction}\label{sec:Introduction}
\noindent
The volatility of renewable energy sources introduces fluctuations in energy generation. 
Demand response (DR) encompasses the management of flexible electricity demand and is expected to help balancing supply and demand while posing economic benefits to electricity consumers~\cite{Zhang.2016}.
We focus on non-dispatchable DR, which is to shift production in time to optimize resource costs based on electricity price signals while ensuring safe operation and meeting product demands.
In this regard, DR requires forecasting algorithms to provide accurate predictions of volatile system parameters like generation, demand, or prices to determine optimal operating decisions~\cite{LennartMerkert.2015}. 

Due to their increasing accuracy, neural network-based forecasting models are superseeding traditional statistical forecasting approaches~\cite{Lago.2021}.
However, their high number of parameters and poor extrapolation abilities can make neural networks prone to adversarial attacks~\cite{Goodfellow.2015}. 
Adversarial attacks are crafted data manipulations that cause a model to produce erroneous outputs while remaining hard to distinguish from benign data. 
In energy systems literature, adversarial attacks are also known as false data injection attacks~\cite{Liu.2011}. 
While the terms adversarial attacks and false data injection attacks are often used interchangeably, the term adversarial attacks originates from machine learning (ML) literature~\cite{Goodfellow.2015} while false data injection attacks were originally introduced in the context of grid state estimation~\cite{Liu.2011}. 
For the sake of clarity, we use the term adversarial attack in this paper. 
Adversarial attacks can cause malicious effects in energy systems and are therefore a major concern regarding future energy systems with interconnected devices and autonomous decision-making algorithms~\cite{JingboHao.2022}. 
Adversarial attacks can manipulate locational marginal prices in energy markets~\cite{Alsharif.2025,Zografopoulos.2023} and compromise state estimation algorithms in the electricity grid~\cite{Tian.2022}. 
Due to the possible negative impacts of adversarial attacks, a plethora of methods for the detection of adversarial attacks has been proposed~\cite{Cui.2020}.

Adversarial attacks on ML models in energy systems have attracted considerable research attention in recent years.
In seminal work, Chen et al.~\cite{Chen.2018} demonstrated the susceptibility of ML models in energy systems to adversarial attacks. 
The authors show  case studies in power quality classification and building load foreacasting and observe significant decreases in ML model performance under adversarial attacks.
Rahman et al.~\cite{M.A.Rahman.2024} review forecasting algorithms in energy systems and their resilience to cyber attacks. 
The authors identify six application areas of forecasting algorithms in energy grids, namely load forecasting, generation forecasting, forecasting in transmission and distribution, forecasting in energy markets, critical event forecasting, and forecasting in attack detection. 
Chen et al.~\cite{Chen.2019} investigate adversarial attacks on load forecasting models in power systems and find that black-box adversarial attacks can have significant negative effects on forecasting performance and power system operations. 
Ruan et al.~\cite{Ruan.2024} explore the vulnerability of photovoltaic (PV) and wind power forecasting algorithms. They also analyze resulting degradations and impacts on optimal power flow problems. 
Their results reveal high economic losses as well as potential instabilities in the power grid as a consequence of manipulated forecasts. 
Tang et al.~\cite{Tang.2021} study adversarial attacks on solar power forecasts and find that gradient-based attack schemes are able to cause strong deviations in the forecasting of solar intensity. 
The authors further show that adversarial training enhances the forecasting  robustness while reducing forecasting accuracy. 
The effect that adversarial training enhances robustness while reducing accuracy was also found by Heinrich et al.~\cite{Heinrich.2024}, who investigate adversarial attacks and adversarial training for wind power forecasts. 
The problem of wind power forecasting was further investigated by Yang et al.~\cite{Yang.2024}, who find that wind forecasting models are especially prone to adversarial attacks in times of high wind speeds. 
Chen et al.~\cite{Chen.2023} propose an adversarial attack on power system inertia forecasting. 
They find that adversarial attacks on inertia forecasting can significantly increase power system operation costs. 
These studies reveal the potential threat of adversarial attacks on forecasting algorithms in energy systems.

While many studies investigate adversarial attacks on power, load, and generation forecasting, the area of manipulated energy price forecasting (EPF) remains rather unexplored~\cite{M.A.Rahman.2024}, albeit being crucial for use cases like industrial DR and many others.
Industrial DR deals with energy-flexible production in the process industries~\cite{Samad.2012}. 
The process industries are a major consumer of electricity and a therefore a major actor in DR. 
The role of the process industries in the electricity market is becoming more significant due to an increase in electrolysis processes and electric heating, as electrification of industrial processes is a major aspect in reducing industrial carbon emissions and increase energy system sustainability~\cite{Scholtz.2024}. 

The economic potential of industrial DR has been studied in a range of research articles. 
For example, Br\'ee et al.~\cite{Bree.2019}  and Otashu and Baldea~\cite{Otashu.2019} find substantial economic potential for the energy-flexible operation of chlor-alkali production processes. 
R\"oben et al.~\cite{Roben.2022} study DR strategies for copper electrolysis, while Sch\"afer~\cite{Schafer.2019} identify economic benefits of industrial DR for aluminum electrolysis. 
Besides electrolysis processes, Yang et al.~\cite{Zhang.2015} as well as Caspari et al.~\cite{Caspari.2019}, amongst others, investigate the DR potential of air separation units. 
Hadera et al.~\cite{HubertHadera.2015} investigate the DR potential of steel production. 
Furthermore, a range of sucessful industrial implementations of DR has been reported~\cite{Samad.2012}. 

DR schedules the internal production based on external data, as DR relies on accurate electricity price forecasting (EPF).  
The dependence on external data makes DR potentially prone to adversarial attacks, as the attacks may take place independently of the internal production scheduling. 
In our previous work~\cite{Cramer.2024}, we show that adversarial attacks can lead to significant economic losses in DR activities. 
Our investigation shows that adversarial attacks on the EPF models have the potential to deteriorate the scheduling decisions in DR, even in cases without explicit knowledge of the underlying EPF and scheduling optimization models. 
The work in~\cite{Cramer.2024} is the first to investigate adversarial attacks in the context of industrial DR as well as arbitrage electricity trading with battery energy storages systems (BESS).

Research on data manipulations in process operations is mainly carried out on the control level, see, e.g.,~\cite{Ye.2025}. 
We refer to the review article by Parker et al.~\cite{Parker.2023} for an overview of cybersecurity aspects of process control. 
In separate lines of work, other authors investigate the impact of adversarial attacks on decision-making frameworks on energy system level or for residential DR. 
For example, adversarial attacks on load forecasting models can lead to pronounced negative impacts on energy system operations in the form of economic losses as well as disruptions and blackouts~\cite{Chen.2019, Chen.2023}. 
Similarly, such effects can also result from adversarial attacks on PV and wind power forecasting~\cite{Ruan.2024}. 
For a discussion of adversarial attacks and defenses in the context of residential DR, the reader is referred to~\cite{Zhang.2024}. 
While these studies investigate the downstream effects of adversarial attacks on DR, an understanding and detailed investigation of the contributing factors to vulnerabilities in industrial DR is still missing. 
As industrial DR must preserve feasibility of the production process at all times, the optimization of production schedules is subject to a set of operational constraints. 
However, it is an open question to what extend different levels of process flexibility influence the vulnerability of DR to adversarial attacks. 

While EPF models are commonly trained on established regression metrics like Mean Squared Error (MSE) and Mean Absolute Error (MAE), these metrics are not necessarily a descriptive metric for the decision quality in DR~\cite{Maciejowska.2026}.
As a consequence, from a demand side perspective, the actual value of an electricity price forecast does not only depend on forecasting accuracy but also on the mathematical properties of the decision problem that the forecasts are utilized for. 
For example, Zareipour et al.~\cite{Zareipour.2009} show that demand side loads in the process industries have different sensitivities to forecasting errors than municipal water-plants. 
Based on these findings, we expect that also the impact of adversarial attacks on industrial DR problems depends on the properties of the different demand side processes.

This article investigates how different levels of process flexibility in industrial DR influence the vulnerability of the coupled system of EPF and DR to adversarial attacks in terms of economic losses. 
In order to enable comparisons beyond single processes, we make use of a generalized process model (GPM)~\cite{Schafer.2020, Germscheid.2022}, which can be adjusted to resemble an abstract representation of a range of industrial processes. 
We systematically vary the process flexibility of the GPM to investigate the impact of adversarial attacks on process scheduling optimization problems.
In addition, we conduct simulations for process configurations that emulate the operational constraints of a range of energy-intensive electrolysis processes.

We progress with our analysis as follows. 
In Section~\ref{sec:attack_scheme}, we introduce the basics of adversarial attacks, DR, and the GPM and outline the different elements of our simulations.
In Section~\ref{sec:num_experiments}, we present and discuss our numerical results. 
In Section~\ref{sec:industrial_cases}, we present the results for adversarial attacks on four illustrative electrolysis processes.
We give the conclusions and outlook of our paper in Section~\ref{sec:Conclusions}.

\section{Attack scheme on industrial demand response}\label{sec:attack_scheme}

\begin{figure*}[h!bt]
	\begin{center}
	  \includegraphics[width=\textwidth, keepaspectratio]{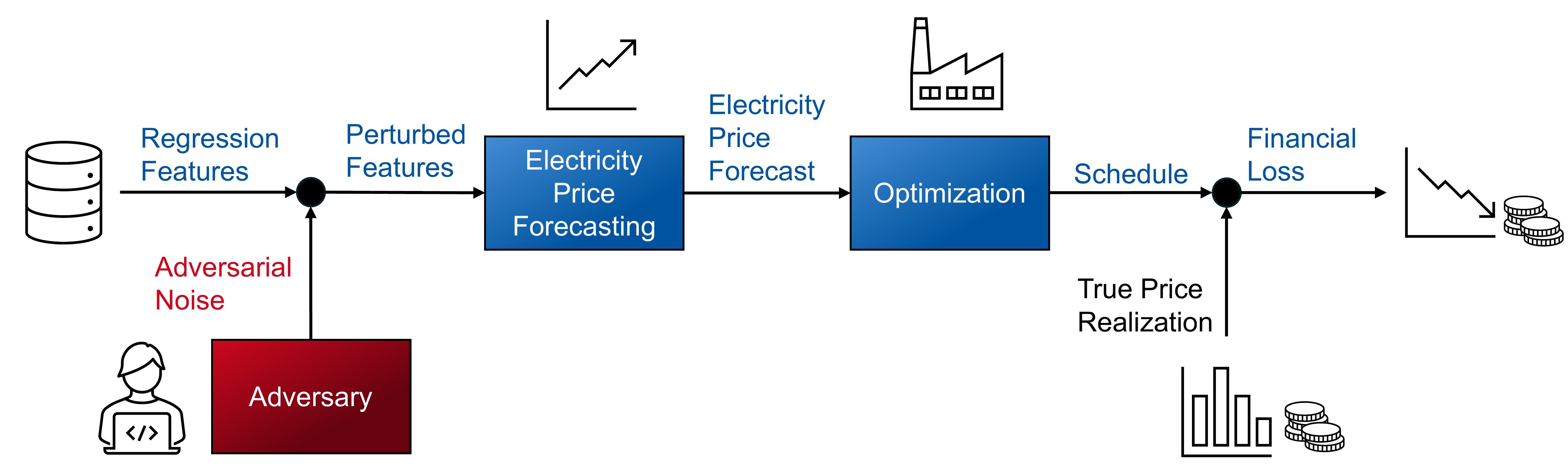}
	  \caption{Overall structure of the adversarial attacks on EPF and DR.}
    \label{fig:DR_structure}
	\end{center}
\end{figure*}

\noindent 
Our investigation consists of three main parts, namely the EPF, the adversarial attack design, and the DR scheduling optimization, see Figure~\ref{fig:DR_structure}. 
In Section~\ref{subsec:epf}, we briefly describe the forecasting problem and the ML model we employ for EPF. 
In Section~\ref{subsec:Attacks}, we present our adversarial attack methodology on the two-step process of energy-price forecasting and DR scheduling optimization.
In Section~\ref{subsec:GPM}, we revisit the generalized process model and describe its emulation of different operational process characteristics of industrial DR.

\subsection{Electricity price forecasting}\label{subsec:epf}

\noindent EPF for day-ahead prices provides crucial market price predictions for decision-making in electricity markets~\cite{Nowotarski.2016}. 
The electricity prices result from an interplay of electricity generation and demand, where the generation is increasingly fluctuating with a rising share of renewable electricity in the grid. 
Consequently, wind and solar generation together with demand have been shown to be key determinants of the price formation in modern day-ahead markets, as they determine the residual loads~\cite{Trebbien.2023}. 
Due to the challenges in EPF introduced by the volatility of renewable generation, a wide range of methods has been proposed for EPF~\cite{Lago.2021}. 

We use a convolutional neural network (CNN) for EPF as these have shown accurate forecasting performance compared to other deep learning models~\cite{Zhang.2020}.
Similar to our previous work~\cite{Cramer.2024} and based on a study by Trebbien et al.~\cite{Trebbien.2023}, we use as input features the day-ahead load forecast $\mathbf{W}^{\mathrm{DA}}_{\mathbf{Load}}$, the day-ahead onshore and offshore wind generation forecasts $\mathbf{W}^{\mathrm{DA}}_{\mathbf{Wind, on}}$ and $\mathbf{W}^{\mathrm{DA}}_{\mathbf{Wind, off}}$, the day-ahead PV forecast $\mathbf{W}^{\mathrm{DA}}_{\mathbf{PV}}$, and the day-ahead price of the day before $\mathbf{P}^{\mathrm{DA}}_{t-1\mathrm{day}}$. 
The model simultaneously predicts the full price vector for all time steps of a day.
This approach aims at better learning the structural relations within the daily price profiles, as this structural relations are crucial for the decision-making in the scheduling optimization problems. 
Further, this structure aligns with the simultaneous determinaton of electricity prices on the German day-ahead market.

We train our model for EPF on the German day-ahead electricity market with electricity price data for 2023 and 2024 obtained from the ENTSO-E transparency platform~\cite{ENTSOEtransparencyplatform.2026}.
We put aside the data for November 2024, which we solely use as test data. From the remaining data, we randomly choose 10\% per epoch as validation data. 
Note that in 2023 and 2024 the day-ahead market used to trade in hourly time intervals while it switched to trading intervals for every quarter of an hour beginning October 2025.

The CNN model consists of two convolutional layers and one linear layer, see Table~\ref{tab: Conv_Model}. The resulting model has 374.000 parameters. 
We train the model with early stopping, an early stopping patience of 25 epochs and a maximum number of 100 epochs.
Further, we use the adam optimizer with a learning rate of $5\cdot10^{-4}$, and a batch size of 32.
The model is trained using \textit{pytorch}~\cite{Paszke.2019}. 
Data pre- and postprocessing is done in \textit{scikit-learn}~\cite{Pedregosa.2011}. 

\begin{table}[ht]
    \centering
    \caption{Structure of the CNN model used for EPF.}
    \begin{tabularx}{\columnwidth}{lXl}
        \toprule
        Layer  & Attributes     & Activation \\
        \midrule
        Convolutional 2D & 16 channels, $7\times3$ & ReLU  \\
        Convolutional 2D & 32 channels, $3\times3$ & ReLU  \\
        Linear & 24         & -  \\ \bottomrule
    \end{tabularx}
    \label{tab: Conv_Model}
\end{table}

\subsection{Adversarial Attacks}\label{subsec:Attacks}

\noindent Adversarial attacks are deliberately crafted data modifications that aim to manipulate the output of ML models. 
At the same time, adversarial attacks should be imperceptible to the human user to remain undetected.
Originally, the investigations of adversarial attacks were concerned with classification problems, see, e.g.,~\cite{Goodfellow.2015, Kurakin.2018}.
However, adversarial attacks can also be applied to regression problems~\cite{KavyaGupta.2021}. 
Adversarial attacks can be categorized into poisoning and evasion attacks, where poisoning attacks are applied during the training phase while evasion attacks manipulate the inputs of already trained ML models~\cite{Biggio.2013}. 
This study is concerned with evasion attacks, i.e., the attacks are applied to models that are already trained, and we use the term adversarial attacks to refer to evasion attacks throughout this work.

The goal of an adversarial attack can be to either drag the model output away from the true target output or to drag the output towards a specific target output~\cite{Kurakin.2018}.
The first is usually referred to as untargeted attack while the latter is referred to as targeted attack. 
For a trained ML model $\mathbf{f}(\mathbf{\cdot})$ with inputs $\mathbf{x}$ and outputs $\mathbf{y}=\mathbf{f}(\mathbf{x})$, adversarial attacks deal with the design of adversarial noise $\Delta\mathbf{x}$ such that the ML model produces a deteriorated output $\tilde{\mathbf{y}}$, i.e., $\tilde{\mathbf{y}} = \mathbf{f}(\mathbf{x} + \Delta\mathbf{x})$. 

In the untargeted case, the attacker aims to maximize the deviation between a true target $\mathbf{y^*}$ and the output produced by the attacked ML model $\tilde{\mathbf{y}} = \mathbf{f}(\mathbf{x} + \Delta\mathbf{x})$, i.e., 

\begin{equation}
  \label{prob: Untargeted Attack} \tag{UA}
  \begin{aligned}
    \underset{\Delta\mathbf{x}}{\max}~\mathcal{L}(\mathbf{x} + \mathbf{\Delta}\mathbf{x},\mathbf{y^*}) &= \sum_{t=1}^{T} \left( y^*_t - \mathbf{f}(\mathbf{x}+\Delta\mathbf{x})_t \right)^2\\
    \mathrm{s.t.}             & \vert\vert \Delta\mathbf{x} \vert\vert_p \leq \varepsilon,
  \end{aligned}
\end{equation}

where the inequality contraint restricts a p-norm of the adversarial noise to a small number $\varepsilon$, i.e., the magnitude of the adversarial noise is restricted such that the perturbations of the input features cannot become arbitrarily large. 

In the targeted case, the attacker aims to minimize the distance between the output of the model and a specified target output $\hat{\mathbf{y}}$, i.e., 
\begin{equation}
  \label{prob: Targeted Attack} \tag{TA}
  \begin{aligned}
    \underset{\Delta\mathbf{x}}{\min}~\mathcal{L}(\mathbf{x} + \mathbf{\Delta}\mathbf{x},\hat{\mathbf{y}}) &= \sum_{t=1}^{T} \left( \hat{y}_t - \mathbf{f}(\mathbf{x}+\Delta\mathbf{x})_t \right)^2\\
    \mathrm{s.t.}             & \vert\vert \Delta\mathbf{x} \vert\vert_p \leq \varepsilon,
  \end{aligned}
\end{equation}

Again, the purpose of the inequality is to restrict a p-norm of the adversarial noise to a small number $\varepsilon$. 

An established algorithm for computing the adversarial noise $\Delta\mathbf{x}$ is the fast gradient sign method (FGSM)~\cite{Goodfellow.2015}. 
FGSM computes the adversarial noise $\Delta\mathbf{x}$ by taking a single step in the direction of the sign of the gradient to solve the optimization problem of the adversarial attack. 
FGSM is known to produce rather spiky adversarial noise which is relatively easy to identify. 
To produce more stealthy attacks, we compute the adversarial noise by means of the basic iterative method (BIM)~\cite{Kurakin.2018}. 
The BIM generates adversarial examples by adding repeated gradient steps, such that it can be seen as an iterative version of FGSM. 
While the iterative updating scheme in BIM makes the attack slightly more expensive, it can produce adversarial examples that are both more powerful and more stealthy than adversarial examples generated by FGSM.

The untargeted attack aims to drag the model output away from the true target $\mathbf{y^*}$. 
However, the true target $\mathbf{y^*}$ is unknown in regression tasks. 
We therefore create a proxy of the true target by using the model output plus a small injection of white-noise $\epsilon \sim \mathcal{N}(0,\sigma^2)$, i.e., $\mathbf{y^*} \approx \mathbf{f}(\mathbf{x}) + \epsilon$. 
As the untargeted attack is a maximization problem, the gradient step needs to be an \emph{ascent} step. 
Consider an initial input $\tilde{\mathbf{x}}^{(0)} = \mathbf{x}$ and a number of iterations $M \in \mathbb{N}$. 
For the untargeted attack, the adversarial input $\tilde{\mathbf{x}}$ is computed iteratively for $ m = 0,1, ...,M-1$ via 

\begin{equation}\label{eq:bim_untargeted}
\tilde{\mathbf{x}}^{(m+1)} = \mathrm{Clip}_{\mathbf{x},\varepsilon} \bigl(\tilde{\mathbf{x}}^{(m)} + \alpha \cdot \mathrm{sign}(\nabla_{\mathbf{x}} \mathcal{L}(\tilde{\mathbf{x}}^{(m)}, \mathbf{f}(\mathbf{x}) + \epsilon)) \bigr),
\end{equation}
with a step size $\alpha > 0$ and a projection
\begin{equation}\label{eq:bim_targeted}
\mathrm{Clip}_{\mathbf{x},\varepsilon}(\tilde{\mathbf{x}}) = \min(\mathbf{x} + \varepsilon, \max(\mathbf{x} - \varepsilon, \tilde{\mathbf{x}}))
\end{equation}

that makes each iteration satisfy the inequality constraint in Problem~\eqref{prob: Untargeted Attack}. 
After the last iteration, we set $\tilde{\mathbf{x}} = \tilde{\mathbf{x}}^{(M)}$.

For the targeted attack, the iteration scheme in Equation~\eqref{eq:bim_untargeted} is adapted to 

\begin{equation}\label{eq:bim_targeted}
\tilde{\mathbf{x}}^{(m+1)} = \mathrm{Clip}_{\mathbf{x},\varepsilon} \bigl(\tilde{\mathbf{x}}^{(m)} - \alpha \cdot \mathrm{sign}(\nabla_{\mathbf{x}} \mathcal{L}(\tilde{\mathbf{x}}^{(m)}, \hat{\mathbf{y}})) \bigr).
\end{equation}

For both the untargeted and the targeted attack, we run the BIM algorithm with $M$ = 10 iterations and set $\alpha = \frac{4 \varepsilon}{3 M}$  based on literature values from image manipulation~\cite{Kurakin.2018}. 
Note that BIM does not actually solve the Problems~\eqref{prob: Untargeted Attack} and~\eqref{prob: Targeted Attack} but takes a number of steps towards the solution of the problem.

To construct a target vector $\hat{\mathbf{y}}$ for the targeted attack, we propose a \textit{mirror} heuristic, that is, we design the attack to drag the EPF model output towards a price profile mirrored around the mean value of the considered day. 
The intuition behind this is to fool the process scheduling into buying electricity at hours where the price is actually high. 
Mathematically, we express the target vector as 

\begin{align}
    \hat{\mathbf{y}}= \mu_{\mathbf{y}} - 3 \cdot(\mathbf{y}-\mu_{\mathbf{y}}),
\end{align}

where $\mu_{\mathbf{y}}$ denotes the mean price of the price prediction of the current day. 
Further, we multiply the profile with a factor of three to amplify the price signal.
As we compute the target vector only based on the predictions of the forecasting model, it can be computed independently of the true prices $\mathbf{y}^*$

As a last step, we remove possible manipulations of PV forecasts during night hours that can be introduced by the BIM. 
As solar generation during night hours is unphysical, it would make the adversarial noise easy to detect.
Further, we set negative values for PV forecasts, offshore wind generation, onshore wind generation, and load to zero, as these are unphysical as well.
By removing these adversarial noise patterns in PV forecasts, we make the attacks more stealthy.

In constrast to our previous paper~\cite{Cramer.2024}, we focus on white-box attacks as our main goal is to understand how adversarial attacks impact the decisions made in the DR optimization problem.
We note that the attacks could also indirectly result from attacks on generation and load forecasts studied in other works~\cite{M.A.Rahman.2024}, as the features of our EPF model are themselves forecasted by ML models, e.g., load and generation forecasts.

\subsection{Generalized Process Model}\label{subsec:GPM}

\noindent The purpose of industrial DR is to flexibly adjust electricity consumption of production processes in order to generate economic benefits from time-varying electricity prices on the electricity markets.
To this end, methods from constrained numerical optimization have proven effective for DR as they provide the possibility to optimize the production schedule while satisfying operational process constraints~\cite{Zhang.2016}.
In order to assess the DR potential of a wide range of industrial processes, different authors have proposed generalized mathematical models of production processes~\cite{Barth.2018,Wanapinit.2021,Schafer.2020,Germscheid.2022}. 
Many energy-intensive processes with the possibility of load shifting and onsite product storage can be seen as ``grid-level storage batteries''~\cite{JoannahI.Otashu.2018}.
The mathematical modeling of such components for DR is a field of active study with early works already starting in the 1980s~\cite{Daryanian.2002,Gellings.1985}.
A wide range of such grid-level storage batteries processes can be modeled with the GPM~\cite{Schafer.2020, Germscheid.2022}.
The GPM consists of general building blocks like storage capacities, ramping constraints, and possible production ranges. 
Sch\"afer et al.~\cite{Schafer.2020} introduce the GPM to study and discuss the interdependencies of economic and sustainability objectives in industrial DR for participation in day-ahead markets. 
Germscheid et al.~\cite{Germscheid.2022} use an extended version of the GPM to investigate the economic potential of industrial DR in intraday markets. 
They further use the model to emulate three industrial electrolysis processes and find that participation in intraday markets can substantially increase the economic potential of industrial DR. 

In constrast to Germscheid et al.~\cite{Germscheid.2022}, we only consider trading on the day-ahead market and not on the intraday market. 
We formulate the GPM as a quasi-steady-state process with an hourly discretization scheme and time steps $t \in {1, ...,T}$, $T=24$ and a duration per time step of $\Delta \mathrm{t}=1\mathrm{h}$.
The flexibility of a process in the GPM is characterized by its nominal power intake $P_{\mathrm{nom}}$, its process oversizing $\theta_{\mathrm{add}}$, its minimal part-load $\theta_{\min}$, its product storage capacity $S$, its ramping limit $R$, its efficiency loss $\zeta$, and its maximum down-time $\tau_{\mathrm{off}}$, see Table~\ref{tab: parameters gpm}. 
Depending on the values of these parameters, there exist three different configurations of the GPM. 
The first formulation resembles an ideal storage-type consumer without efficiency losses and without the possibility of intermittent production shutdown. 
A second configuration considers the possibility of production shutdowns, and a third considers efficiency losses in the production dependent on the operation point. 
In the following, we present the modeling equations of the three different process configurations.

\begin{table}
\centering
\caption{Parameters of the generalized process model. The parameters $\theta_{\mathrm{add}}$, $\theta_{\min}$, $S$, $R$, and $\zeta$ are given as a multiple of the nominal power intake $P_{\mathrm{nom}}$. Table adapted from~\cite{Germscheid.2022}.}
\label{tab: parameters gpm}
\begin{tabularx}{\columnwidth}{lXl}
\toprule
Parameter & Description & Unit \\
\midrule
$P_{\mathrm{nom}}$ & Nominal power intake & [MW] \\
$\theta_{\mathrm{add}}$ & Process oversizing & [-] \\
$\theta_{\min}$ & Minimal part-load & [-] \\
$S$ & Product storage capacity & [h] \\
$R$ & Ramping limit per hour & [1/h] \\
$\zeta$ & Efficiency loss at minimal part-load  & [-] \\
& compared to $P_{\mathrm{nom}}$ & \\
$\tau_{\mathrm{off}}$ & Maximum consecutive down-time & [h] \\
\bottomrule
\end{tabularx}
\end{table}

The ideal storage-type consumer is represented by Eqs.~\eqref{eq:power_uptake}-\eqref{eq:cyclic_storage}, where $p_{t}$ represents the power intake at time $t$. The variable $r_t$ represents the effective production rate at time t, and $l_t$ represents the fill level of the storage tank at time t. 
Eq.~\eqref{eq:power_uptake} restricts the maximum and minimum power intake of the process. 
Eq.~\eqref{eq:overall_demand} restricts the total production per day to the equivalent of a production amount in steady-state operation. 
As the ideal storage type customer does not account for efficiency losses, the effective production rate equals the nominal power intake, see Eq.~\eqref{eq:no_efficiency_losses}.
Eq.~\eqref{eq:ramping_limits} enforces the ramping limits of the process. 
Eq.~\eqref{eq:storage_level} defines the change in the storage tank level by the difference of a fixed production amount per hour and the effective production rate. 
The upper and lower bound of the storage level are given by Eq.~\eqref{eq:max_storage_capacity}. 
The initial and final storage tank level are constrained to be at 50\% capacity in order to enforce cyclic behavior, see Eq.~\eqref{eq:cyclic_storage}. 
Note that Eq.~\eqref{eq:storage_level} and Eq.~\eqref{eq:cyclic_storage} implicitly enforce Eq.~\eqref{eq:no_efficiency_losses} but are added for the potential case of a process without storage capacities~\cite{Germscheid.2022}. 

\begin{align}
    P_{\mathrm{nom}}\theta_{\min} 
    &\le p_{t} 
    \le P_{\mathrm{nom}}\left(1+\theta_{\mathrm{add}}\right),
    && \forall t \in \{1,\dots,T\}, 
    \tag{1}\label{eq:power_uptake} \\[4pt]
    P_{\mathrm{nom}} T 
    &= \sum_{t=1}^{T} r_{t}, 
    \tag{2}\label{eq:overall_demand} \\[4pt]
    r_{t} 
    &= p_{t},
    && \forall t \in \{1,\dots,T\}, 
    \tag{3}\label{eq:no_efficiency_losses} \\[4pt]
    -P_{\mathrm{nom}} R 
    &\le \frac{p_{t+1}-p_{t}}{\Delta t} 
    \le P_{\mathrm{nom}} R,
    && \forall t \in \{1,\dots,T-1\}, 
    \tag{4}\label{eq:ramping_limits} \\[4pt]
    l_{t+1} 
    &= l_{t} + (r_{t}-P_{\mathrm{nom}})\Delta t,
    && \forall t \in \{1,\dots,T\}, 
    \tag{5}\label{eq:storage_level} \\[4pt]
    0 
    &\le l_{t} \le P_{\mathrm{nom}} S,
    && \forall t \in \{1,\dots,T\}, 
    \tag{6}\label{eq:max_storage_capacity} \\[4pt]
    l_{1} 
    &= l_{T+1} = 0.5 P_{\mathrm{nom}} S, 
    \tag{7}\label{eq:cyclic_storage}
\end{align}

For a process with the possibility of process shutdowns during hours with peak prices, Eqs. \eqref{eq:max_shutdown_duration}-\eqref{eq:symmetry_breaking} can be added to describe the shutdown of the process. 
Herein, $\tau_{off}$ denotes the maximum allowed number of consecutive shutdown hours and $\bar{\tau}_{off}= \tau_{off}/h$ denotes the corresponding index.
Eq.~\eqref{eq:max_shutdown_duration} enforces that $\tau_{off}$ is not violated and Eqs. \eqref{eq:transition},\eqref{eq:symmetry_breaking} describe the binary logic that allows the shutdown possibilities. 
Therein, the binary variable $\gamma_t$ describes whether the process is being shutdown or not. 
The binary variables $z_{\mathrm{on},t}$ and $z_{\mathrm{off},t}$ indicate a transition from off to on and from on to off, respectively. 
Eqs. \eqref{eq:lb_ramping_shutdown}-\eqref{eq:power_uptake_shutdown} replace the constraints Eqs. \eqref{eq:power_uptake} and \eqref{eq:ramping_limits} to combine the process ramping and operating limits with the shutdown opportunities.

\begin{align}
    \gamma_{t+{\bar{\tau}_{\mathrm{off}}}} &\ge z_{\mathrm{off},t}, \notag \\
    & \forall t \in \{1,\ldots,T - \bar{\tau}_{\mathrm{off}}\}, \tag{13}\label{eq:max_shutdown_duration}\\[4pt]
    z_{\mathrm{on},t} - z_{\mathrm{off},t} &= \gamma_{t+1} - \gamma_t,  \notag \\
    & \forall t \in \{1,\ldots, T - 1\}, 
    \tag{14}\label{eq:transition}\\[4pt]
    z_{\mathrm{on},t} + z_{\mathrm{off},t} &\le 1,  \notag \\
    & \forall t \in \{1,\ldots, T \}, 
    \tag{15}\label{eq:symmetry_breaking}\\[4pt]
    - \gamma_{t+1} P_{\mathrm{nom}} R 
    - \bigl(1 - \gamma_{t+1}\bigr)\frac{P_{\mathrm{nom}}\theta_{\min}}{\Delta t}
    &\le \frac{p_{t+1} - p_{t}}{\Delta t}, \notag \\
    & \forall t \in  \{1,\ldots,T-1\}, \tag{16}\label{eq:lb_ramping_shutdown}\\[4pt]
    \gamma_{t} P_{\mathrm{nom}} R 
    + \bigl(1 - \gamma_{t}\bigr)\frac{P_{\mathrm{nom}}\theta_{\min}}{\Delta t}
    & \ge \frac{p_{t+1} - p_{t}}{\Delta t}, \notag \\
    & \forall t \in \{1,\ldots,T-1\}, \tag{17}\label{eq:ub_ramping_shutdown}\\[4pt]
    \gamma_{t} P_{\mathrm{nom}}\theta_{\min} \le p_{s,t} 
    &\le 
    \gamma_{t} P_{\mathrm{nom}}(1 + \theta_{\mathrm{add}}), \notag \\
    & \forall t \in \{1,\ldots,T\}. \tag{18}\label{eq:power_uptake_shutdown}
\end{align}

In order to model efficiency losses, we consider different formulations dependent on the characteristics of the process. We give further details on the modeling of efficiency losses in the description of  the case studies. 

The scheduling objective is to minimize the electricity costs of the process, that is, $\sum_{t=1}^{T}P_{t}^{\mathrm{DA}}  p_t$. 
Therein, the price vector $\mathbf{P}^{\mathrm{DA}}$ is the output of the EPF model, i.e., $\mathbf{P}^{\mathrm{DA}}= \mathbf{f}(\mathbf{x})$. 
To compute the actual costs under the true price realizations, we compute the actual costs as $\sum_{t=1}^{T}P_{t}^{\mathrm{DA}^*}  p_t$, with true price realizations $\mathbf{P}^{\mathrm{DA}^*}$.

To evaluate a potential interplay between process flexibility and vulnerability to adversarial attacks, we systematically vary the process oversizing and therefore the flexibility of an illustrative flexible process. 
This illustrative process is characterized by a nominal power intake $P_{\mathrm{nom}}=1.0~\mathrm{MW}$, a minimal part load $\theta_{\mathrm{min}}=0.5$, a product storage capacity $S=3~\mathrm{h}$, a ramping limit per hour $R=1.00$, an efficiency loss $\zeta=0$, and a maximum consecutive down time $\tau_{\mathrm{off}}=0~\mathrm{h}$. 
All models are implemented in Pyomo~\cite{Bynum.2021} and solved via Gurobi~\cite{GurobiOptimization.}.

\section{Impact analysis of adversarial attacks on industrial demand response}\label{sec:num_experiments}

\noindent We investigate the impact of adversarial attacks on the two-step process of EPF and scheduling optimization. 
In Section~\ref{subsec:impact_forecast}, we investigate the impact of adversarial attacks on the output of the electricity price forecasting model. 
In Section~\ref{subsec:single_paras}, we discuss how these perturbed outputs impact DR scheduling optimization problems. 
In Section~\ref{subsec:optimal_basis}, we analyze the interplay of financial losses in DR under adversarial attacks to changes in the optimal basis of DR scheduling problems. 
In Section~\ref{subsec:relative_savings}, we further analyze the interplay between process flexibility and the impact of adversarial attacks on DR. 
Finally, in Section~\ref{subsec:seasonal_effects}, we discuss seasonal effects in the impact of adversarial attacks on DR.
All numerical experiments are carried out on a 4th Generation Intel(R) Xeon(R) Platinum 8468 Processor with 2.10 GHz on a Rocky Linux 9 operating system.

\subsection{Impact of adversarial attacks on electricity price forecast}\label{subsec:impact_forecast}

\noindent
We start our analysis with a brief presentation of the training results of the EPF model. 
The model reaches its lowest validation loss after 14 epochs.
We use early stopping to stop further training after a patience of 25 epochs and restore the model weights from epoch 14. 
With these weights, the EPF model reaches an MSE of 0.11 on both the training set and on the test month of November 2024. 
We use this model configuration in all remaining analyses of this work.

We now turn to a comparison of the impact of the adversarial attacks on the output of the EPF model. 
Our goal herein is to compare the magnitude of errors induced by the different attack heuristics, i.e., targeted and untargeted attacks. 
To this end, we compare the MSE and MAE of price forecasts under the influence of adversarial attacks in relation to the price forecasts in the case without attacks. 
Note that we only compare forecasted profiles with each other and do not compare the forecasts to the true realizations. 
This is because we want to analyze the deviations induced by the adversarial attacks and not the absolute predictive performance of the models.

\begin{figure}[h!bt]
	\begin{center}
	  \includegraphics[width=0.7\textwidth, keepaspectratio]{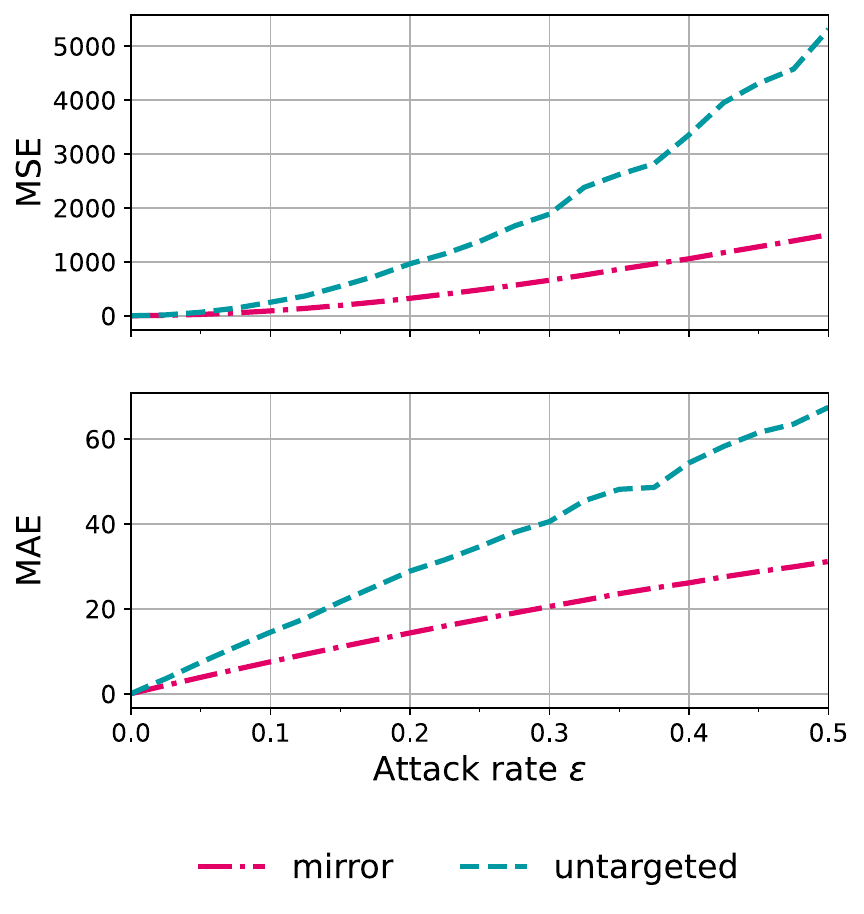}
	  \caption{Cumulated error metrics between the benign forecasts and the attacked forecasts for November 2025.}
    \label{fig:profits}
	\end{center}
\end{figure}

Our results show that both attacks induce significant deviations from the original forecasting outputs, see Figure~\ref{fig:profits}. 
With both attack heuristics, both the MSE and the MAE grow with increasing attack rates. 
Further, the untargeted attack induces two to three times larger deviations than the mirror attack in both metrics. 

\begin{figure}[h!bt]
	\begin{center}
	  \includegraphics[width=0.7\textwidth, keepaspectratio]{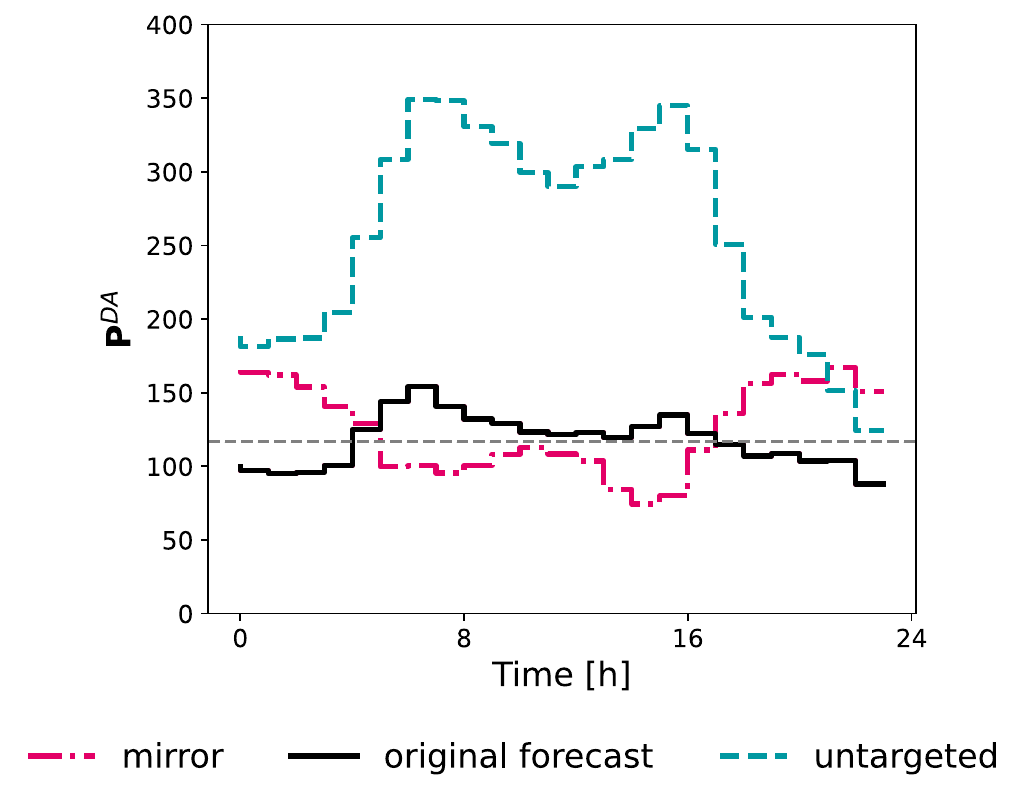}
	  \caption{Comparison of electricity profiles resulting from the targeted and untargeted attack heuristics for November 29, 2024. In both cases, we apply an extreme attack rate of $\varepsilon=1.0$ and compare it to the case without attack, i.e., $\varepsilon=0.0$. Gray dashed lines indicate the mean of the forecast of the day.}
    \label{fig:price_results}
	\end{center}
\end{figure}

We explain the difference in severity with the different design heuristics of the two attacks. 
The mirror attack aims at mirroring the price profile around the mean price of the day. 
Therefore, the mean of the perturbed price profiles remains close to the unperturbed price profiles. 
In contrast, the untargeted attack can increase the prediction error by shifting the whole price profile to lower or higher price regimes, inducing significantly larger error metrics than the mirror attack. 
We illustrate this behavior by showing two exemplary manipulated forecasts under an extreme attack rate of $\varepsilon=1.0$, i.e., in the order of one standard deviation, to showcase the different results of the attacks, see Figure~\ref{fig:price_results}. 
While the untargeted attack produces a price profile that is completely shifted to a lower price regime, the mirror attack produces a price profile that approximately mirrors the original price profile around the mean price. 

\begin{figure*}[h!bt]
	\centering
	\begin{subfigure}{\textwidth}
		\centering
		\includegraphics[width=\textwidth]{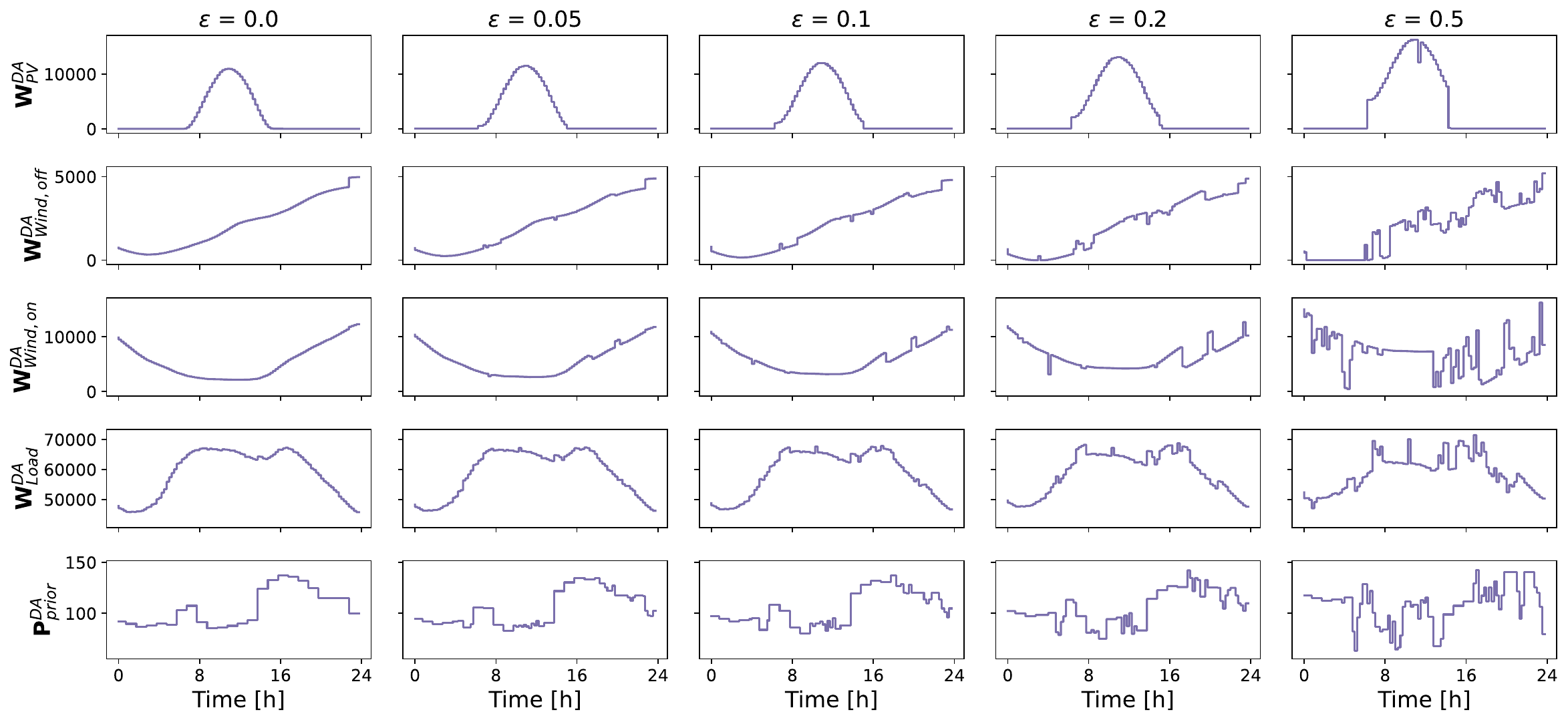}
		\caption{}
		\label{fig:perturbations_mirror}
	\end{subfigure}
	\begin{subfigure}{\textwidth}
		\centering
		\includegraphics[width=\textwidth]{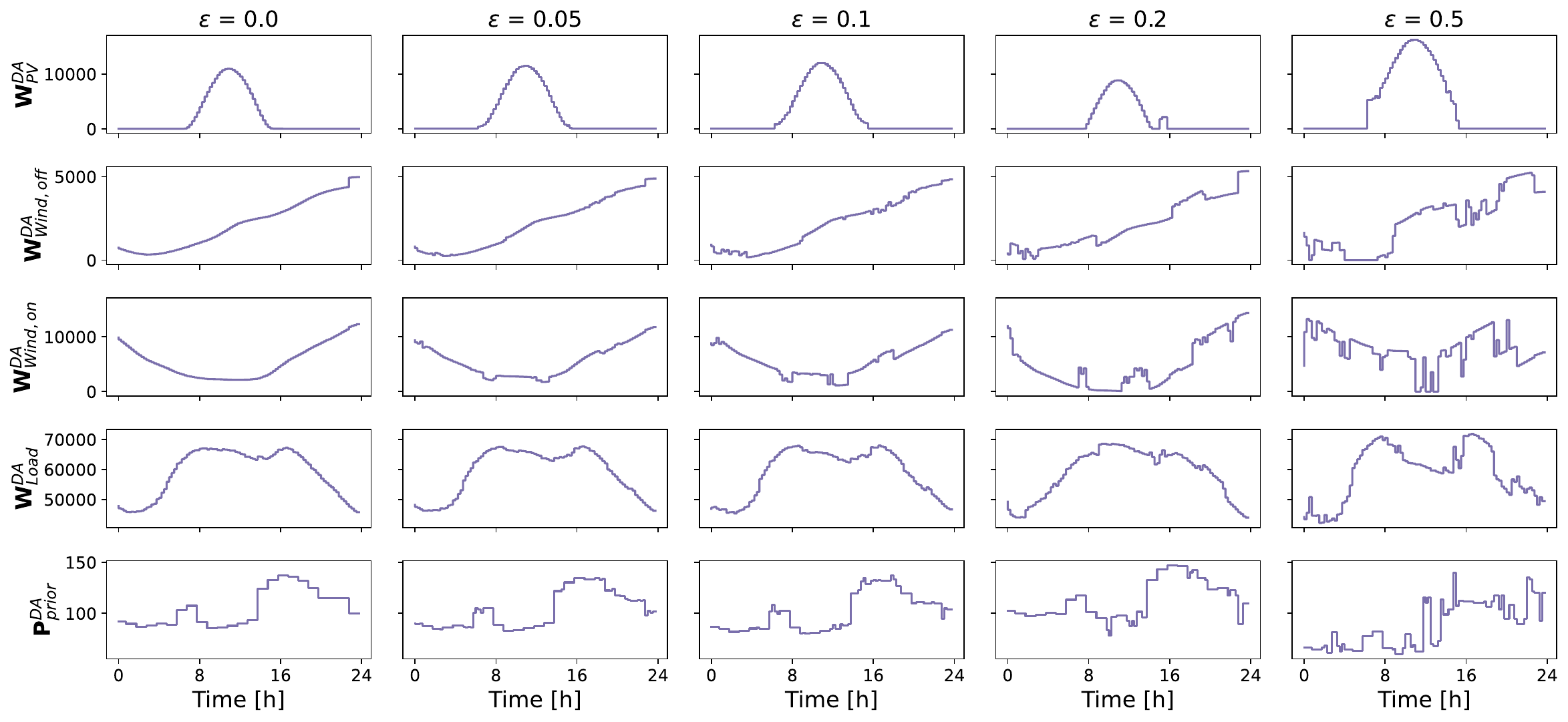}
		\caption{}
		\label{fig:perturbations_untargeted}
	\end{subfigure}
	\caption{Comparison of feature perturbations resulting from the targeted and untargeted attack heuristics for November 29, 2024. Note that we have filtered unphysical perturbations. (a) Result for the mirror attack. (b) Result for the untargeted attack.}
	\label{fig:perturbations}
\end{figure*}

While the adversarial attacks aim at pronounced negative impact in decision-making, the attacks should remain stealthy. 
We visualize the adversarial perturbations in the features of the forecasting model in Figure~\ref{fig:perturbations}. 
The figure shows the perturbed features for both attack heuristics and for different attack rates for November 29, 2024. 
Note that we filtered the perturbations for the features in such a way that all negative values for the PV forecast, the offshore wind production forecast, the onshore wind generation forecast, am=nd the load are replaced by a value of zero. 
Negative values for these features are physically impossible and corresponding perturbations are therefore trivial to detect.

Both mirror and untargeted attack induce the largest perturbations in the day-ahead PV forecast $\mathbf{W}^{\text{DA}}_{\textbf{PV}}$ and in the day-ahead wind onshore forecast $\mathbf{W}^{\text{DA}}_{\textbf{Wind, on}}$. 
We note that the untargeted attack tends to cause negative values for the PV forecast such that the filtering of negative values can remove a large portion of the perturbations stemming from the untargeted attack in some cases. 

When comparing the perturbations for different attack rates we see that the attacks become increasingly easy to detect with larger attack rates. 
For attack rates exceeding an attack rate of 0.1, the attacked features significantly deviate from the original data, such that these large deviations are comparably easy to identify by the human user. 
Therefore, we mark an attack rate of 0.1 as the maximum attack rate under which the attacks can still remain unrecognized. 


For completeness, we also show the perturbations stemming from an attack rate of 0.5. 
These perturbations are easy to detect due to their jumps, for example in the wind onshore forecasts $\mathbf{W}^{\text{DA}}_{\textbf{Wind, on}}$.

In conclusion, the untargeted attack produces more pronounced forecasting errors while the mirror attack leads to price profiles that are closer to the original output in terms of MSE and MAE. 
For both attack heuristics, we refer to an attack rate of 0.1 as the maximum attack rate under which the attacks can still remain stealthy.

\subsection{Variation of process flexibility}\label{subsec:single_paras}
\noindent  We investigate the costs of process operation under adversarial attacks by means of the relative costs compared to steady-state process operation. 
We vary the flexibility of the process by a structured variation of the process oversizing $\theta_{\mathrm{add}}$. 
For all investigated process configurations, we compute the process electricity costs under attack rates ranging from 0.0 to 0.5 discretized by steps of size 0.025 for the mirror attack heuristic as well as the untargeted attack heuristic. 
We simulate price forecasts and process optimization for the test data from November 2024 and report results as the cumulated costs over the complete month.

\begin{figure}[h!bt]
	\begin{center}
	  \includegraphics[width=0.7\textwidth, keepaspectratio]{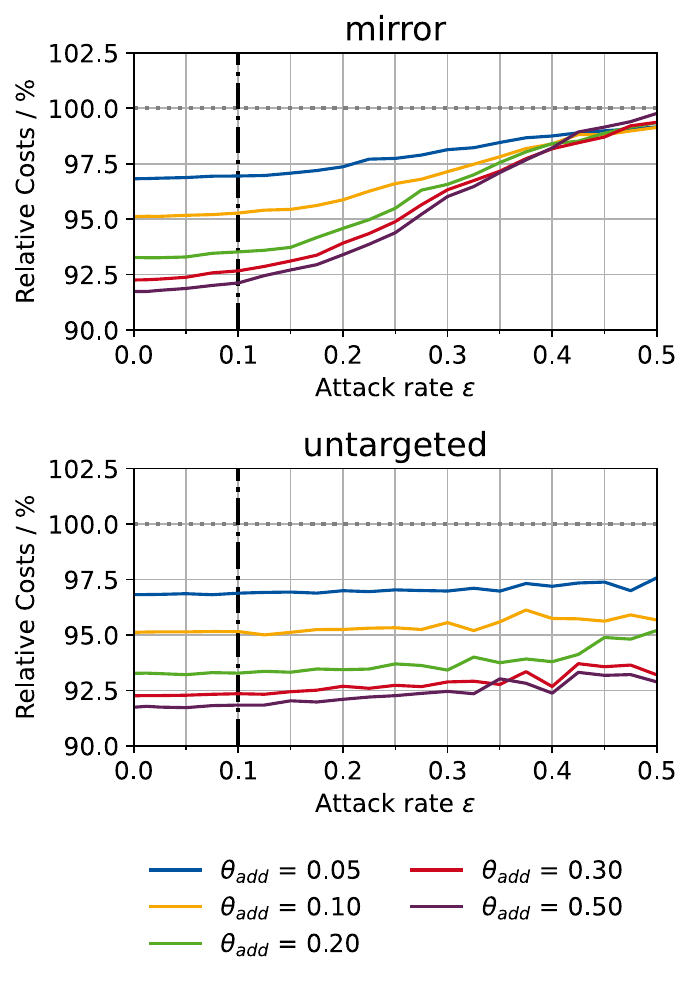}
	  \caption{
      Relative costs of processes with different process overzising. costs of 100\% are equal to the costs in case of steady-state operation. The black dashed line indicates the maximal attack rate that we still refer to as stealthy.
      The process is characterized by a nominal power intake $P_{\mathrm{nom}}=1.0~\mathrm{MW}$, a minimal part load $\theta_{\mathrm{min}}=0.5$, a product storage capacity $S=3~\mathrm{h}$, a ramping limit per hour $R=1.00$, an efficiency loss $\zeta=0$, and a maximum consecutive down time $\tau_{\mathrm{off}}=0~\mathrm{h}$. }
    \label{fig:relative_cost_WB_theta_add_nov_final}
	\end{center}
\end{figure}

We simulate the process electricity costs compared to steady-state operation for a range of $\theta_{\mathrm{add}}$ from 0.05 to 0.50. 
The results without attacks, i.e., with an attack rate of 0.0, show that an increase in process oversizing leads to increased process flexibility.
Consequently, the relative costs compared to steady-state operation are the lowest with a process oversizing of 50\%, see Figure~\ref{fig:relative_cost_WB_theta_add_nov_final}. 
With an oversizing of 5\%, the process allows for only 3.2\% savings, while with 50\% oversizing, up to 8.3\% of electricity costs can be saved. 

For all considered process configurations, the electricity costs increase with increasing attack rates $\varepsilon$. 
In the case of the mirror attack, the costs increase moderately up to the threshold at the attack rate of 0.1. 
For higher attack rates, the process costs approach the costs of steady-state operation.
At the maximum atack rate of 0.5, the processes with the largest oversizing equal steady-state process costs.
Also, the effect of process flexibility is inverted, i.e., higher flexibility leads to higher losses for attack rates above 0.35. 
The more flexible the process is, the more its operation schedule is being adapted to the perturbed electricity price forecasts. 

For the untargeted attack, the impact of adversarial attacks on the process schedule is significantly lower than for the mirror attack. 
Even with the maximum attack rate of $\varepsilon$=0.5, the process costs with DR are significantly lower than the steady-state costs, i.e., performing DR remains economically beneficial even under severely deteriorated electricity price forecasts. 
For attack rates up to 0.1, the costs are almost invariant to the adversarial attacks.
This is the case for all considered process configurations. 

The impact of the two different adversarial attack heuristics on decision-making is inverted to their impact on EPF. 
While the untargeted attack leads to more pronounced errors in EPF than the mirror attack, it is vice versa for the decision-making, where the mirror attack induces significantly higher costs than the untargeted attack.

\subsection{Changes in the optimal basis of the optimization problem}\label{subsec:optimal_basis}
\noindent

We explain the difference in the impacts of untargeted and mirror attack by the following intuition. 
The GPM in its considered configuration without shutdown opportunities constitutes a linear program (LP).
The feasible set of an LP consititutes a convex polytope.
The electricity price forecast determines the parameters in the objective function of the DR optimization problem.
Therefore, changes in electricity price predictions resulting from adversarial attacks only influence the objective function of the LP scheduling problem while the feasible region of the LP remains unaffected by the adversarial attacks. 
If an LP has a unique solution, this solution lies at a vertex of the feasible polytope of the LP~\cite{Nocedal.2006}.
In this sense, the adversarial attacks need to shift the optimum of the LP to a different vertex of the feasible polytope of the problem.

Changes of the cost vector within a critical region affect the optimal objective value but not the optimal solution point. Only changes beyond a critical region change the optimal solution of the scheduling optimization problem as the changes need to be sufficient to shift the optimal solution to a different vertex of the feasible polytope.
The untargeted attack tends to increase the forecasting error by shifting the complete electricity price vector to higher or lower price regimes. 
As a simple scaling of the objective function of an LP does not change its optimal solution, the shift of the price profile as induced by the untargeted attack induces less significant changes in the solution of the LP. 
In contrast, the mirror attack induces changes in the orientation of the electricity price vector. Consequently, the mirror attack leads to higher economic losses compared to the untargeted attack when the scheduling decisions are evaluated under the true price realizations.

To further illustrate the impact of adversarial attacks on the solution of the LP, we analyze the changes in the optimal basis of the process scheduling problem that are induced by the adversarial attacks. 
Based on our intuition, we expect that the costs in the scheduling problem are positively correlated with the changes in the optimal basis of the problem. 
That is, an increase in process costs comes with an increase in changes in the optimal basis of the optimal scheduling solution. 

We define the changes in the optimal basis as the share of exchanged variables in the basis at the optimal solution in percent. 
We note that the formulation of the GPM can exhibit redundant constraints and variables. 
Redundant variables or constraints can have side effects on the changes in the optimal basis, for example multiple variables changing their basis status at once. 
To avoid such side effects, we analyze the changes in the optimal basis after an intial presolve routine of the solver Gurobi~\cite{GurobiOptimization.}. 
The presolved model consists of 18 variables and 41 constraints.

\begin{figure}[h!bt]
	\begin{center}
	  \includegraphics[width=0.7\textwidth, keepaspectratio]{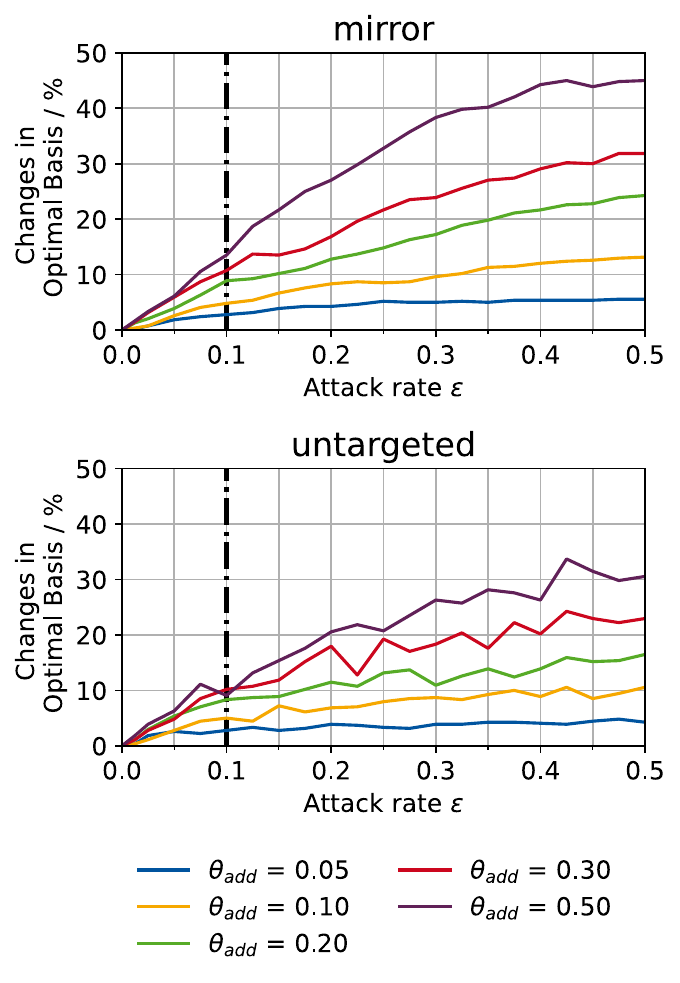}
	  \caption{Changes in the optimal basis of the process scheduling optimization problem compared to the case without adversarial attack for processes with different process oversizing. The process is characterized by a nominal power intake $P_{\mathrm{nom}}=1.0~\mathrm{MW}$, a minimal part load $\theta_{\mathrm{min}}=0.5$, a product storage capacity $S=3~\mathrm{h}$, a ramping limit per hour $R=1.00$, an efficiency loss $\zeta=0$, and a maximum consecutive down time $\tau_{\mathrm{off}}=0~\mathrm{h}$. The black dashed line indicates the maximal attack rate that we still refer to as stealthy.}
    \label{fig:basis_changes_WB_theta_add_nov_final}
	\end{center}
\end{figure}

We observe that the changes in the optimal basis are indeed positively correlated with the attack rate, see Figure~\ref{fig:basis_changes_WB_theta_add_nov_final}. 
Therein, we plot the changes in the optimal basis for attack rates ranging from 0.0 to 0.5, i.e., for the same simulations as shown in Figure~\ref{fig:relative_cost_WB_theta_add_nov_final}. 
We refer to the basis that stems from a price forecast without adverarial attack as the default solution and optimal basis, that is, we do not assume perfect foresight of electricity prices in the default solution.

We make three main observations with regards to the changes in the optimal basis. First, the changes in the optimal basis increase with increasing attack rates $\varepsilon$, both for the mirror attack and for the untargeted attack. 
Second, we observe that for processes with a large process flexibility, the changes in the optimal basis is larger than for processes with less process flexibility, in this case, with less process oversizing. 
For example, the changes in the optimal basis are approx. 45~\% for an attack rate of 0.5 and an oversizing of 0.50, while it is only 5~\% for the same attack rate and an oversizing of 0.05.
Third, we observe that the mirror attack induces more changes in the optimal basis than the untargeted attack which is in line with the economic effects of the adversarial attacks as shown before in Figure~\ref{fig:relative_cost_WB_theta_add_nov_final}. 

The observation that the untargeted attack induces large errors in the electricity price forecasts but less errors in the decision-making is in line with the finding that MSE and MAE alone are not necessarily descriptive measures of the resulting decision quality, see~\cite{Zareipour.2009}. 
Instead, we outline that the specific geometric properties of the LP optimization problem lead to the result that the mirror attack induces significantly larger deviations in the decision-making outcome than the untargeted attack. 
Thereby, we can conclude that in order to design effective attacks on industrial DR, the sensitvities of the optimization problem need to be taken into account, as we have already implicitly done in the design of he mirror attack. 

\subsection{Comparison of relative cost savings for different process configurations}\label{subsec:relative_savings}

So far, we have observed that higher process flexibility increases both the absolute economic losses under adversarial attacks and also the changes in the optimal basis under adversarial attacks. 
Further, we have observed that flexible processes have higher potential for cost savings compared to steady-state process operation, cf. Figure~\ref{fig:relative_cost_WB_theta_add_nov_final}. 
Based on these findings, we now aim to investigate whether the increases in costs stemming from the adversarial attacks are also larger in relative terms, i.e., how much of the cost savings in DR is retained under adversarial attacks. 
We define the relative cost savings under an attack rate $\varepsilon$ as 

\begin{equation}\label{eq:relative_savings}
\mathrm{Relative Cost Savings (\varepsilon)} = \frac{{\mathrm{costs_{steady-state}} - \mathrm{costs} (\varepsilon) }}{\mathrm{costs_{steady-state}} - \mathrm{costs}_{\varepsilon=0.0} },
\end{equation}

where $\mathrm{costs_{steady-state}}$ denotes the costs for steady-state process operation, $\mathrm{costs} (\varepsilon)$ denotes the costs under an attack rate $\varepsilon$, and $\mathrm{costs}_{\varepsilon=0.0}$ denotes the costs for DR without adversarial attacks, i.e., with an attack rate of 0.0.

\begin{figure}[h!bt]
	\begin{center}
	  \includegraphics[width=0.7\textwidth, keepaspectratio]{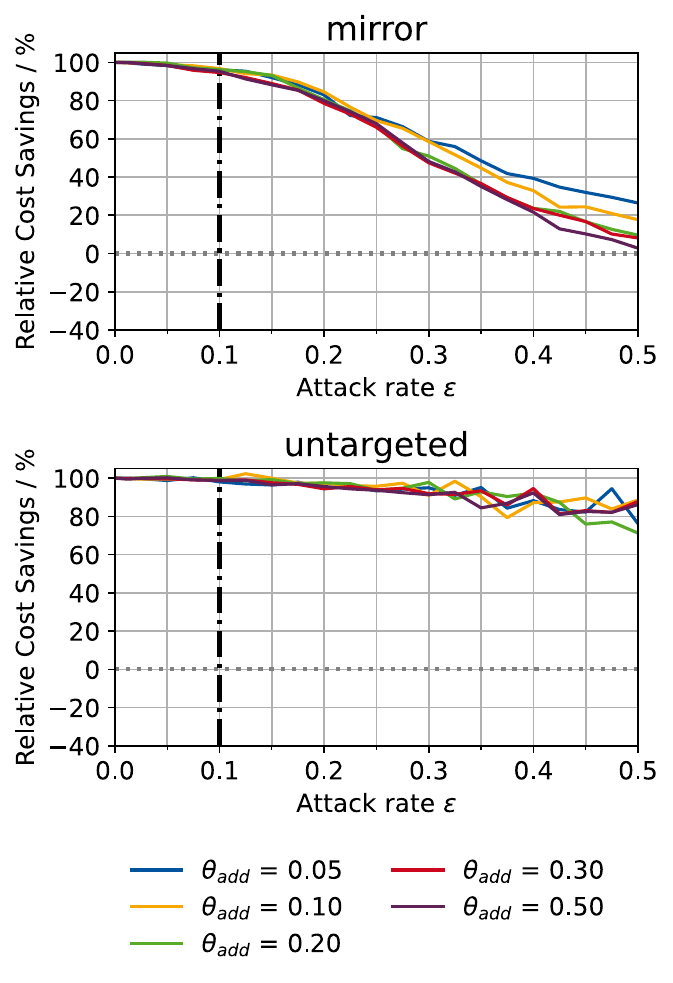}
	  \caption{Relative cost savings compared to steady-state process operation with different process oversizings. The process is characterized by a nominal power intake $P_{\mathrm{nom}}=1.0~\mathrm{MW}$, a minimal part load $\theta_{\mathrm{min}}=0.5$, a product storage capacity $S=3~\mathrm{h}$, a ramping limit per hour $R=1.00$, an efficiency loss $\zeta=0$, and a maximum consecutive down time $\tau_{\mathrm{off}}=0~\mathrm{h}$. The black dashed line indicates the maximal attack rate that we still refer to as stealthy.}
    \label{fig:savings_WB_theta_add_nov_final}
	\end{center}
\end{figure}

The results show that the impact of adversarial attacks on the relative cost savings is almost invariant to process flexibility, see Figure~\ref{fig:savings_WB_theta_add_nov_final}, where all differnet levels of process flexibility show a similar decrease in relative cost savings under adversarial attacks. 
For high attack rates beyond 0.25, the decrease in savings begins to show a distinct behavior, with more flexible processes showing slightly higher losses than less flexible processes, a result that can be similarly observed for the absolute decrease in costs, cf. Figure~\ref{fig:savings_WB_theta_add_nov_final}. 

For the untargeted attack, all processes retain 90 to 95 \% of savings up to an attack rate of 0.1.
Notably, for the untargeted attack, all processes retain more than 70\% of relative cost savings even in the case of the largest attack rates, while with the mirror attack, the same attack rate completely erodes the financial benefits of DR. 

Overall, the results show that DR optimization problems are largely robust against adversarial attacks of reasonable magnitude. 
While an increase in process flexibility leads to a larger absolute decrease in cost savings, the relative savings in DR compared to steady-state operation are largely invariant to the process flexibility.
Further, while mirror attacks lead to less pronounced errors in electricity price forecasting, they lead to significantly larger errors in decision-making.

\subsection{Analysis of seasonal effects}\label{subsec:seasonal_effects}
\begin{figure}[h!bt]
	\begin{center}
	  \includegraphics[width=0.7\textwidth, keepaspectratio]{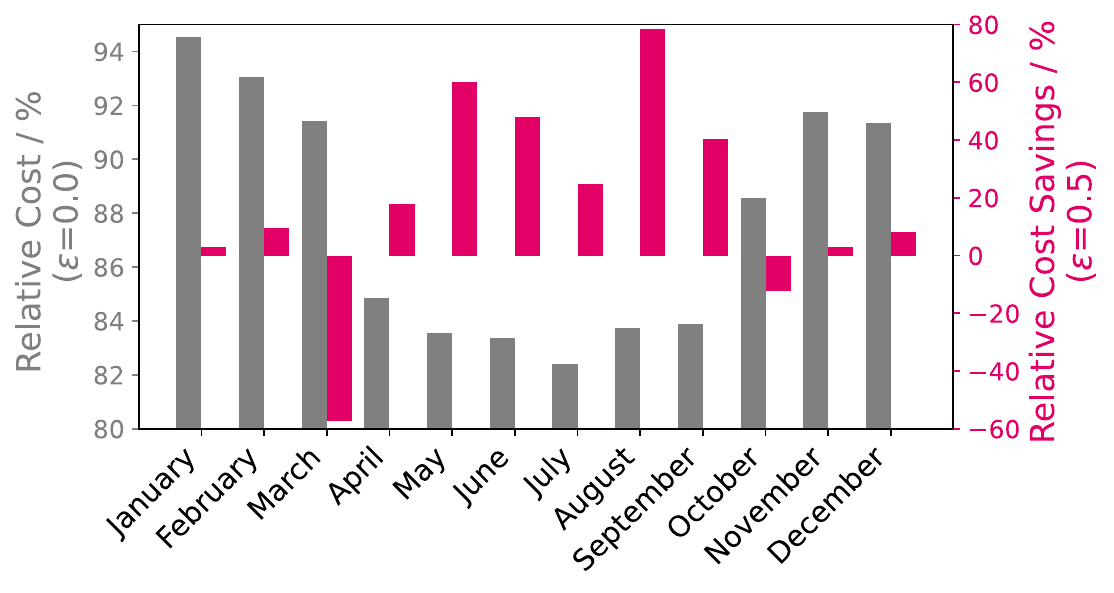}
	  \caption{Comparison of MSE on the test set and the relative costs savings for November 2024. The relative cost savings are computed with the mirror attack. The process is characterized by a nominal power intake $P_{\mathrm{nom}}=1.0~\mathrm{MW}$, an oversizing of $\theta_{add}=0.5$, a minimal part load $\theta_{\mathrm{min}}=0.5$, a product storage capacity $S=3~\mathrm{h}$, a ramping limit per hour $R=1.00$, an efficiency loss $\zeta=0$, and a maximum consecutive down time $\tau_{\mathrm{off}}=0~\mathrm{h}$.}
    \label{fig:bar_plot_CostvsSavings_theta_add_compare_months_final}
	\end{center}
\end{figure}

\noindent
We have so far anaylzed simulations for test data from November 2024. As electricity price profiles follow seasonal trends, we investigate the impacts of adversarial attacks over the course of the year 2024. 
For this comparison, we choose the process model with the largest process oversizing of 50 \%. 
For each month, we retrain the price forecasting model with the corresponsing month as test set. 
Then, we first compute the relative costs compared to steady-state, that is, the same quantity as in Figure~\ref{fig:relative_cost_WB_theta_add_nov_final}. 
For an attack rate of 0.0, this quantity serves as a measure of the economic potential of DR for the corresponding month, see the grey bars in Figure~\ref{fig:bar_plot_CostvsSavings_theta_add_compare_months_final}. 
Second, we compute the relative cost savings, that is, the same quantity as in Figure~\ref{fig:savings_WB_theta_add_nov_final}, for the mirror attack. 
For an attack rate of 0.5, this quantity serves as a measure for the potential maximum impact of the adversarial attacks on the costs of the process, see the pink bars in Figure~\ref{fig:bar_plot_CostvsSavings_theta_add_compare_months_final}.

The comparison of these two quantities reveals a clear seasonal trend. 
As the potential for DR is higher in summer months, the relative costs are lower during summer months and higher during winter months. 
Consequently, the financial benefits of DR in summer are larger than in winter.

In contrast, the relative cost savings under attack show the opposite trend. 
During summer months, the financial benefits of DR remain even under the highest attack rate of 0.5, while for winter months, the relative cost savings reach values around zero under an attack rate of 0.5.

The shape and orientation of the price profile during summer months is significantly influenced by the high production of solar energy. 
Therefore, the mirror attack struggles to invert the orientation of the price profile such that the DR scheduling problem finds solutions that perform significantly worse than under the unperturbed price predictions.

These findings reveal that the impact of adversarial attacks has a strong seasonal component. During summer, DR offers high financial benefit and is relatively robust against adversarial attacks. 
In contrast, during winter months, the potential of DR is lower and also, the impact of adversarial attacks is more pronouced. 

\section{Impact of adversarial attacks on four illustrative process case studies}\label{sec:industrial_cases}

\noindent
We apply our adversarial attack scheme to four GPM configurations that emulate the flexibility of different electrolysis processes. In Section~\ref{subsec:description_case_studies}, we first analyze the different properties and potentials for DR of the four considered processes. Then, we analyze the impact of the adversarial attacks in Section~\ref{subsec:impact_case_studies}. Lastly, we investigate the influence of non-convexities on the impact of the adversarial attacks in Section~\ref{subsec:non_convexity}.

\subsection{Process case studies}\label{subsec:description_case_studies}
\noindent
We consider four electrolysis processes, namely, an aluminum electrolysis process~\cite{Schafer.2019}, a copper electrolysis process~\cite{Roben.2022}, a chlor-alkali electrolysis process~\cite{Bree.2019}, and a PEM water electrolysis process~\cite{Mucci.2023}. 
The parametrizations of the considered processes are given in Table~\ref{tab: paras case studies}. 

\begin{table}[h!]
\centering
\caption{Parameters of the four process case studies. The parameters $\theta_{\mathrm{add}}$, $\theta_{\min}$, $S$, $R$, and $\zeta$ are given as a multiple of the nominal power intake $P_{\mathrm{nom}}$. The parameters for the first three electrolysis processes are taken from~\cite{Germscheid.2022}. Details on the modeling of the PEM process are presented in Section~\ref{subsec:description_case_studies}. For details on the efficiency modeling of the chlor-alkali-process and the aluminum electrolysis process, see~\cite{Germscheid.2022}.}
\label{tab: paras case studies}
\begin{tabularx}{\columnwidth}{lXXXX}
\toprule
 &
 &
Chlor‑ &
 &
PEM \\
 &Copper & Alkali &Aluminum & water \\
\midrule
$P_{\mathrm{nom}}$      & 14.97 MW & 2.34 MW & 90 MW   & 100 MW   \\
$\theta_{\mathrm{add}}$ & 0.105    & 0.13    & 0.25    & 0.25     \\
$\theta_{\mathrm{min}}$ & 0        & 0.42    & 0.75    & 0.2625   \\
$S$                      & $\infty$ h & 3 h   & 24 h   & 23.2 h   \\
$R$                      & 4.0/h    & 1.42/h  & 4.0/h   & 1.25/h   \\
$\zeta$                  & 0.0      & *       & 0.01    & *        \\
$\tau_{\mathrm{off}}$    & (0 h)    & (0 h)   & (0 h)   & (24 h)   \\
\bottomrule
\end{tabularx}
\end{table}

For the first three electrolysis processes, the parametrizations are taken from~\cite{Germscheid.2022}. 
For the PEM water electrolysis process, we derive the parameters from a paper by Mucci et al.~\cite{Mucci.2023}.
We use the optimization results from Table A.9 therein, which are computed for volatile electricity price scenarios. 
The value of the hydrogen vessel volume is $5430 \mathrm{m^3}$.
We determine the maximum storage capacity for a maximum storage pressure of 140 bar. 
Therefore, the maximum mass of stored hydrogen $M_{\mathrm{H}_2}(t)$ is 
\begin{align}
M_{\mathrm{H}_2}(t)
\approx V \cdot 0.073~\frac{\mathrm{kg}}{\mathrm{m}^3~\mathrm{bar}}
\cdot \left(140~\mathrm{bar}-75~\mathrm{bar}\right).
\end{align}

With a nominal production rate of $1.11 ~\frac{\mathrm{t}}{h}$, we get a storage capacity of 23.2 hours for our GPM. 
We do not model the methanol plant but instead assume that it requires a constant outlet flow rate of hydrogen at the nominal production rate of $1.11 ~\frac{\mathrm{t}}{h}$. 
We add the possibility of process shutdowns but with an unrestricted duration.
For the efficiency losses, we fit a piecewise linear function with help of scikit-learn~\cite{Pedregosa.2011} consisting of two linear pieces. 
For more information on the ,linearization the reader is referred to the original article~\cite{Mucci.2023}. 
As in the original article, we set the nominal power intake to 100 MW and the minimal part-load to 0.2625. 
We set the ramping limit to 1.25 times the nominal power intake, which means that the process can be ramped in its full operating range within a single time step.

The four considered processes exhibit different mathematical properties. 
All four processes have different definitions of efficiency losses.
Further, the PEM water electrolysis process allows for temporary shutdowns, and consequently, the process model is a mixed-integer linear program (MILP) and therefore non-convex.
Due to their different levels of flexibility, the four processes exhibit different potentials for financial savings in DR. 
We show the electricity costs and savings of the processes in the test month of November 2024 in Table~\ref{tab: savings industrial}. 
We compute the costs and savings using forecasts from our EPF model, i.e., we do not assume perfect price foresight here.
Due to their different nominal power intake, the processes show different electricity costs in steady-state operation and DR.
With potential savings of about 11 \%, the PEM water electrolysis process shows the highest economic potential for DR, while the aluminum electrolysis process only achieves about 4 \% savings. 
Overall, the four processes have different levels of flexibility, different mathematical properties, and different economic potential for DR.

\begin{table*}[h!]
\centering
\caption{Cost savings for the four investigated processes. All costs are cumulated over the test month November 2024.}
\label{tab: savings industrial}
\begin{tabularx}{\textwidth}{Xrrrr}
\toprule
Electrolysis process & costs Steady-State & costs DR & Savings & Relative Savings
\\
\midrule
Copper  & 1.233.000 \texteuro & 1.152.000 \texteuro & 81.000 \texteuro & 6.6 \% \\
Chlor-Alkali  & 193.000 \texteuro & 184.000 \texteuro & 9.000 \texteuro & 4.7 \% \\
Aluminum  & 7.412.000 \texteuro & 7.107.000 \texteuro & 305.000 \texteuro & 4.1\% \\
PEM water  & 8.235.000 \texteuro & 7.339.000 \texteuro & 896.000 \texteuro & 10.9 \% \\
\bottomrule
\end{tabularx}
\end{table*}

\subsection{Economic impact of adversarial attacks on DR}\label{subsec:impact_case_studies}
\begin{figure}[h!bt]
	\begin{center}
	  \includegraphics[width=0.7\textwidth, keepaspectratio]{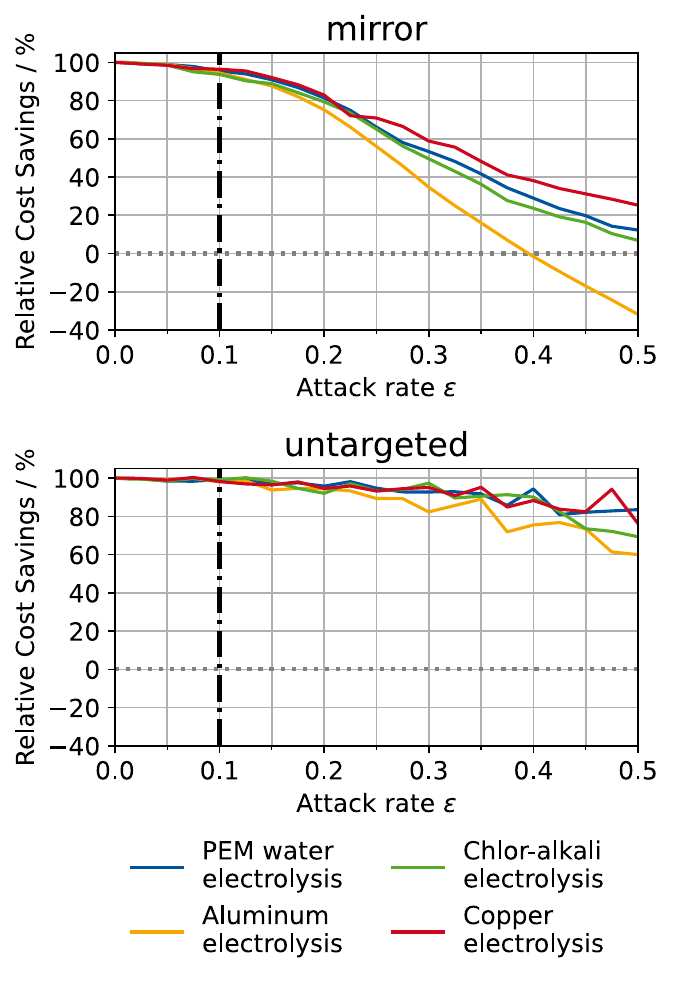}
	  \caption{Relative cost savings compared to steady-state process operation for different electrolysis processes under adversarial attacks. The black dashed line indicates the maximal attack rate that we still refer to as stealthy.}
    \label{fig:savings_WB_industrial_nov_final}
	\end{center}
\end{figure}

\noindent
Along the lines of the analysis in Section~\ref{sec:num_experiments}, we investigate whether DR retains relative savings compared to steady-state operation under the influence of adversarial attacks, cf. Eq.~\eqref{eq:relative_savings}. 
The results for the four considered processes are shown in Figure~\ref{fig:savings_WB_industrial_nov_final}. Similar to the results before, it can be seen that the mirror attack induces more pronounced losses than the untargeted attack.
For the untargeted attack, all processes retain almost 100~\% of their cost savings under an atttack rate of 0.1 and more than 60 \% cost savings even under the largest attack rate of $\varepsilon$=0.5. 
For the mirror attack, all considered processes retain more than 90 \% savings for attack rates of $\varepsilon \leq 0.1$. 
For larger attack rates, the aluminum process shows a slightly steeper decrease in savings, turning savings into losses for attack rates larger than 0.375.  In that case, the adversarial attacks make the DR result in financial losses compared to steady-state production.

We show the absolute increases in electricity costs for different attack rates of the mirror attack in Table~\ref{tab:savings_industrial_under_attack}. 
For a rather small attack rate of 0.05, the costs of the chlor-alkali process with a nominal power intake of 2.34 MW only increase by 100~\texteuro, while for the PEM process with a nominal power intake of 100 MW, the costs already increase by 10.100~\texteuro. 
For an attack rate of 0.1, which we refer to as the maximal stealthy attack rate, the increase in costs for the chlor-alkali process is only 600~\texteuro, while for the PEM process, it is already 40.900~\texteuro. 
These results show that while the adversarial attacks have only small relative impact on the electricity costs, these effects can have large absolute impact if the electricity costs of a process are considerably large, as is the case for the PEM process. 
For the maximal attack rate of 0.5, we would even have an increase in costs for the PEM electrolyzer of alomst 800.000~\texteuro per month. 
However, we note that these attack rates are hardly feasible due to the significant perturbations in the forecasting features.

\begin{table}[h!]
\centering
\caption{Increase in electricity costs for different attack rates of the mirror attack. All costs are cumulated over the test month November 2024.}
\label{tab:savings_industrial_under_attack}
\begin{tabularx}{\columnwidth}{Xrrr}
\toprule
Electrolysis process & $\varepsilon=0.05$ &
$\varepsilon=0.1$ & $\varepsilon=0.5$
\\
\midrule
Copper & 1.200 \texteuro & 2.900 \texteuro & 60.400 \texteuro \\
Chlor-Alkali & 100 \texteuro & 600 \texteuro & 8.000 \texteuro\\
Aluminum & 3.500 \texteuro & 16.800 \texteuro & 401.500 \texteuro \\
PEM water & 10.100 \texteuro & 40.900 \texteuro & 785.500 \texteuro \\
\bottomrule
\end{tabularx}
\end{table}

The overall trends in the relation of cost savings and attack rates remain the same as for the illustrative process configurations in Section~\ref{sec:num_experiments}.
For resonable attack rates, the adversarial attack do not erase the economic benefits from DR compared to steady-state process operation. However, the increase in electricity costs can cumulate to a large amount if the process has a large nominal power intake and therefore high electricity costs.

\subsection{Effects of non-convexity of the scheduling problem}\label{subsec:non_convexity}
\noindent
In contrast to the other scheduling problems, the PEM water electrolysis model constitutes an MILP. 
The integer variables of the MILP make the problem non-convex. 
Therefore, the feasible set of an MILP is not a convex polytope as it is the case for LPs, see the discussion in Section~\ref{subsec:optimal_basis}.
To gain insights into the changes that the adversarial attacks induce in the solution of the MILP scheduling problem, we conduct an analysis similar to the analysis of the LP's optimal basis variables. 
In this regard, we count the number of changes in the binary variables at the optimal solution under adversarial attacks. 

\begin{figure}[h!bt]
	\begin{center}
	  \includegraphics[width=0.7\textwidth, keepaspectratio]{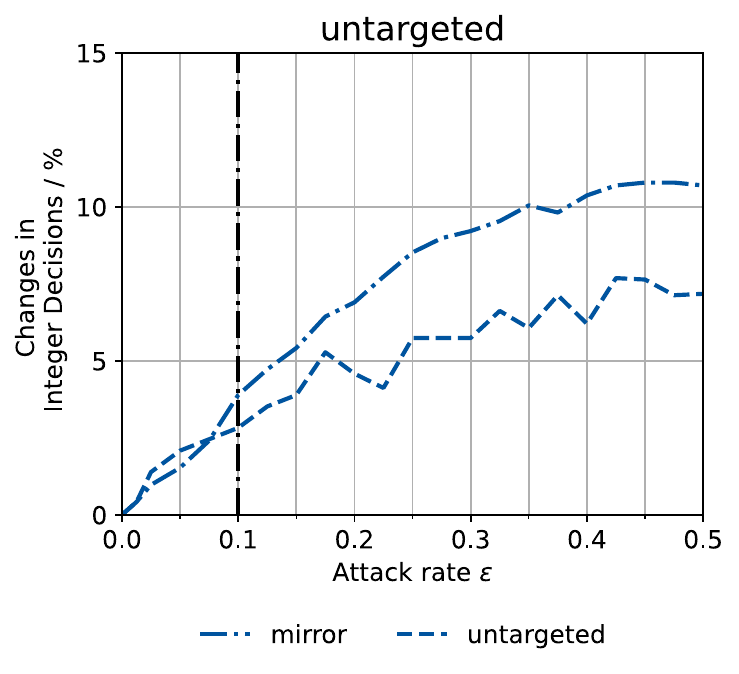}
	  \caption{Changes in the integer decisions of the MILP scheduling problem of the PEM water electrolysis process compared to the case without adversarial attack. The plot shows average values for November 2024. The black dashed line indicates the maximal attack rate that we still refer to as stealthy.}
    \label{fig:integers_changes_WB_industrial_nov_final}
	\end{center}
\end{figure}

We show the results in Figure~\ref{fig:integers_changes_WB_industrial_nov_final}. 
The trend in the number of changes of integer variables behaves similar to the trend in the continuous variables of the LP problems. 
That is, the number of changes increases with larger attack rates. 
For the mirror attack, more than 10 \% of the integer variables change their value under adversarial attacks, while for the untargeted attack, only about 7 \% of the integer decision change under the largest attack rate of 0.5. 

The Figure shows the changes as the average of changes per day as an average value over the month November 2024. 
We observe that while some days show multiple changes in the integer decisions, some days do not show any changes in the integer decisions under adversarial attacks. 
We note that we compare the number of changes in the integer variables for the original model formulation without presolve routines. 
Consequently, some of the integer variables in the MILP model might be redundant or forced to a specific value by some of the constraints.
In contrast, in the analysis in Section~\ref{subsec:optimal_basis}, we had analyzed a presolved version of the model formulation.

Overall, the trend in the changes of integer variables in the MILP scheduling problem is similar to the trend in the continuous variables of the LP scheduling problems. Therefore, we do not find a particular influence of the nonconvexity of the problem on the vulnerability to adversarial attacks. 

\section{Conclusions}\label{sec:Conclusions}
\noindent
We investigate the potential vulnerability of industrial DR to adversarial attacks.
To this end, we apply evasion attacks to deteriorate the predictions of electricity price forecasting models.
We analyze the economic effects of scheduling decisions parametrized by deteriorated electricity price forecasts and compare these economic effects for different levels of process flexibility and different process configurations that resemble a range of electrolysis processes. 

We find that for all investigated process examples, DR retains more than 90 \% cost savings compared to steady-state production under stealthy adversarial attacks. To erode the financial advantages of DR, the adversarial attacks in our work need to be of such a large magnitude that they are easy to detect by the human user.
While processes with an increased process flexibility have potentially higher absolute financial losses under adversarial attacks, they behave similarly to less flexible processes in terms of the impact of adversarial attacks on the relative relative cost advantages gained from DR compared to steady-state operation.
In addition, we find that large errors in forecasting accuracy do not necessarily lead to large economic losses in the DR scheduling problems. 
Consequently, we identify targeted attack heuristics as more potent than untargeted attack heuristics, as the targeted attacks are designed with the properties of the scheduling problem in mind.

Our results show that while DR relies on externally provided market data and forecasts, adversarial attacks have only limited impact on the decisions made in DR, while the forecasting models themselves show pronounced forecasting errors under adversarial attacks.
Further, the impact of adversarial attacks is larger in winter months than in summer months.

We note that all adversarial attacks in this work are white-box attacks in terms of the electricity price forecasting model. 
In practical scenarios, it is difficult to obtain explicit knowledge of the model weights from the perspective of the adversary. 
Therefore, the attacker will often have to rely on black-box attacks~\cite{Cramer.2024}, which are generally less potent than white-box attacks. 

While we have made extensive analyses in terms of different attack designs and process configurations, we see several lines for future work. 
First, we have only investigated adversarial attacks in the day-ahead market.
Therefore, future work should investigate further markets, particularly intraday markets. 
Second, we deem automated anomaly detection methods for adversarial attacks promising to detect adversarial attacks, especially for large attack rates. 
Third, the discrepancy between results for the untargeted and the targeted attack heuristics suggest that more sophisticated targeted attack designs might be capable of inducing more significant financial losses than the rather simple mirror heuristic applied in this work.

\section*{Data availability}
\noindent The training data for the electricity price forecasting models are
publicly available on the Entso-E Transparency Platform (https://
transparency.entsoe.eu/, ENTSO-E Transparency Platform (2026)). This
work uses the day-ahead residual load forecasts, i.e., wind power
generation, solar power generation, and load, and the day-ahead electricity
prices for the BZN|DE-LU bidding zone in 2023 and 2024. Other
research data or code will be made available upon request.

\section*{Acknowledgements}

\noindent Funded by the Deutsche Forschungsgemeinschaft (DFG, German Research Foundation) – 546226379.
This work was also performed as part of the Helmholtz School for Data Science in Life, Earth and Energy (HDS-LEE). Simulations were performed with computing resources granted by RWTH Aachen University.

We sincerely thank Alexander Mitsos for his valuable feedback on the manuscript.

\section*{Declaration of competing interest}
The authors declare that they have no known competing financial interests or personal relationships that could have appeared to influence the work reported in this paper.
\section*{CRediT authorship contribution statement}
\noindent \textbf{Clemens Kortmann}: Conceptualization, Data curation, Formal analysis, Methodology, Software, Visualization, Writing - Original draft.
\textbf{Eike Cramer}: Conceptualization, Funding acquisition, Methodology, Supervision, Writing - Review and Editing.\\

\section*{Nomenclature}
\noindent

\begin{tabularx}{\columnwidth}{lX}
    \hline
    Symbol                                              & Description      \\               
    \hline
    $\Delta\mathbf{x}$                                  & Adversarial noise                 \\
    $\epsilon$                                          & White noise                       \\
    $\sigma^2$                                          & White noise variance              \\
    $\varepsilon$                                       & Attack rate                       \\
    $\mathbf{f}( \cdot )$                               & Machine learning model / EPF model\\
    $\mathbf{x}$                                        & Inputs                            \\
    $\tilde{\mathbf{x}}$                                & Perturbed inputs                  \\
    $\mathbf{y}$                                        & Outputs\\
    $y_i$                                               & i-th dimension of outputs         \\
    $\mathbf{y}^*$                                      & True realization/label            \\
    $\hat{\mathbf{y}}$                                  & Target outputs                    \\
    $\tilde{\mathbf{y}}$                                & Perturbed outputs                 \\
    $\mathcal{L}(\mathbf{x},\mathbf{y})$                & Loss function                     \\
    $\nabla_{\mathbf{x}} \mathcal{L}(\mathbf{x},\cdot)$ & Loss gradient w.r.t. inputs       \\
    $\alpha$                                            & step size in BIM attack           \\
    M                                                   & number of iterations in BIM attack\\
    $\mu_{\mathbf{y}}$                                  & mean of the output $\mathbf{y}$   \\
    $\mathbf{P}^{\mathrm{DA}}$                          & Day-ahead electricity prices      \\
    $\mathbf{P}^{\mathrm{DA}}_{t-1\mathrm{day}}$        & Day-ahead price on previous day   \\
    $P^{\mathrm{DA}}_{t}$                               & Day-ahead price at t-th hour      \\
    $\mathbf{W}^{\mathrm{DA}}_{\mathbf{PV}}$            & Day-ahead PV forecast \\
    $\mathbf{W}^{\mathrm{DA}}_{\mathbf{Wind, on}}$    & Day-ahead wind onshore forecast \\
    $\mathbf{W}^{\mathrm{DA}}_{\mathbf{Wind, off}}$   & Day-ahead wind offshore forecast \\
    $\mathbf{W}^{\mathrm{DA}}_{\mathbf{Load}}$        & Day-ahead load forecast \\ 
    $P_{nom}$                                         & Nominal Power intake\\
    $\theta_{\mathrm{add}}$                           & Process oversizing\\
    $\theta_{\mathrm{min}}$                           & minimal part-load\\
    $S$                                               & Product storage capabity\\
    $R$                                               & Ramping limit per hour \\
    $\zeta$                                           & Efficiency loss at minimal part-load compared to $P_{nom}$ \\
    $\tau_{\mathrm{off}}$                             & Maximum consecutive down-time \\
    $t$                                               & time index\\
    $p_t$                                             & power intake at time t\\
    $r_t$                                             & effective power at time t\\ 
    $T$                                               & number of time steps/hours per day (24)\\
    $l_t$                                             & storage level at time t  \\
    $z_{\mathrm{off},t}$                              & variable indicating transition from production to shutdown\\
    $z_{\mathrm{on},t}$                               & variable indicating transition from shutdown to production\\
    $\gamma_{t}$                                      &  variable indicating whether the process is active at time t\\
    $\Delta \mathrm{t}$                               & Time interval  \\

    \hline

\end{tabularx}

\appendix



    \bibliographystyle{elsarticle-num}
    \bibliography{References.bib}

\begin{thebibliography}{10}
\expandafter\ifx\csname url\endcsname\relax
  \def\url#1{\texttt{#1}}\fi
\expandafter\ifx\csname urlprefix\endcsname\relax\def\urlprefix{URL }\fi
\expandafter\ifx\csname href\endcsname\relax
  \def\href#1#2{#2} \def\path#1{#1}\fi

\bibitem{Zhang.2016}
Q.~Zhang, I.~E. Grossmann, Enterprise-wide optimization for industrial demand side management: Fundamentals, advances, and perspectives, Chemical Engineering Research and Design 116 (2016) 114--131.
\newblock \href {https://doi.org/10.1016/j.cherd.2016.10.006} {\path{doi:10.1016/j.cherd.2016.10.006}}.

\bibitem{LennartMerkert.2015}
{Lennart Merkert}, {Iiro Harjunkoski}, {Alf Isaksson}, {Simo S{\"a}ynevirta}, {Antti Saarela}, {Guido Sand}, \href{https://www.sciencedirect.com/science/article/pii/S0098135414001768}{Scheduling and energy -- industrial challenges and opportunities}, Computers {\&} Chemical Engineering 72 (2015) 183--198.
\newblock \href {https://doi.org/10.1016/j.compchemeng.2014.05.024} {\path{doi:10.1016/j.compchemeng.2014.05.024}}.
\newline\urlprefix\url{https://www.sciencedirect.com/science/article/pii/S0098135414001768}

\bibitem{Lago.2021}
J.~Lago, G.~Marcjasz, B.~de~Schutter, R.~Weron, Forecasting day-ahead electricity prices: A review of state-of-the-art algorithms, best practices and an open-access benchmark, Applied Energy 293 (2021) 116983.
\newblock \href {https://doi.org/10.1016/j.apenergy.2021.116983} {\path{doi:10.1016/j.apenergy.2021.116983}}.

\bibitem{Goodfellow.2015}
I.~Goodfellow, J.~Shlens, C.~Szegedy, Explaining and harnessing adversarial examples, in: 3rd International Conference on Learning Representations, ICLR 2015 - Conference Track Proceedings, 2015, pp. 1--11.

\bibitem{Liu.2011}
Y.~Liu, P.~Ning, M.~K. Reiter, False data injection attacks against state estimation in electric power grids, ACM Trans. Inf. Syst. Secur. 14~(1) (2011).
\newblock \href {https://doi.org/10.1145/1952982.1952995} {\path{doi:10.1145/1952982.1952995}}.

\bibitem{JingboHao.2022}
{Jingbo Hao}, {Yang Tao}, \href{https://www.sciencedirect.com/science/article/pii/S2352484721011707}{Adversarial attacks on deep learning models in smart grids}, Energy Reports 8 (2022) 123--129.
\newblock \href {https://doi.org/10.1016/j.egyr.2021.11.026} {\path{doi:10.1016/j.egyr.2021.11.026}}.
\newline\urlprefix\url{https://www.sciencedirect.com/science/article/pii/S2352484721011707}

\bibitem{Alsharif.2025}
G.~O. Alsharif, C.~Anagnostopoulos, A.~K. Marnerides, Energy market manipulation via false-data injection attacks: A review, IEEE Access 13 (2025) 42559--42573.
\newblock \href {https://doi.org/10.1109/ACCESS.2025.3548914} {\path{doi:10.1109/ACCESS.2025.3548914}}.

\bibitem{Zografopoulos.2023}
I.~Zografopoulos, N.~D. Hatziargyriou, C.~Konstantinou, Distributed energy resources cybersecurity outlook: Vulnerabilities, attacks, impacts, and mitigations, IEEE Systems Journal 17~(4) (2023) 6695--6709.
\newblock \href {https://doi.org/10.1109/JSYST.2023.3305757} {\path{doi:10.1109/JSYST.2023.3305757}}.

\bibitem{Tian.2022}
J.~Tian, B.~Wang, J.~Li, C.~Konstantinou, Adversarial attack and defense methods for neural network based state estimation in smart grid, IET Renewable Power Generation 16~(16) (2022) 3507--3518.
\newblock \href {https://doi.org/10.1049/rpg2.12334} {\path{doi:10.1049/rpg2.12334}}.

\bibitem{Cui.2020}
L.~Cui, Y.~Qu, L.~Gao, G.~Xie, S.~Yu, Detecting false data attacks using machine learning techniques in smart grid: A survey, Journal of Network and Computer Applications 170 (2020) 102808.
\newblock \href {https://doi.org/10.1016/j.jnca.2020.102808} {\path{doi:10.1016/j.jnca.2020.102808}}.

\bibitem{Chen.2018}
Y.~Chen, Y.~Tan, D.~Deka, Is machine learning in power systems vulnerable?, in: 2018 IEEE International Conference on Communications, Control, and Computing Technologies for Smart Grids (SmartGridComm), 2018, pp. 1--6.
\newblock \href {https://doi.org/10.1109/SmartGridComm.2018.8587547} {\path{doi:10.1109/SmartGridComm.2018.8587547}}.

\bibitem{M.A.Rahman.2024}
{M.A. Rahman}, {Md. Rashidul Islam}, {Md. Alamgir Hossain}, {M.S. Rana}, {M.J. Hossain}, {Evan MacA. Gray}, \href{https://www.sciencedirect.com/science/article/pii/S0952197624009436}{Resiliency of forecasting methods in different application areas of smart grids: A review and future prospects}, Engineering Applications of Artificial Intelligence 135 (2024) 108785.
\newblock \href {https://doi.org/10.1016/j.engappai.2024.108785} {\path{doi:10.1016/j.engappai.2024.108785}}.
\newline\urlprefix\url{https://www.sciencedirect.com/science/article/pii/S0952197624009436}

\bibitem{Chen.2019}
Y.~Chen, Y.~Tan, B.~Zhang, Exploiting vulnerabilities of load forecasting through adversarial attacks, in: Proceedings of the Tenth ACM International Conference on Future Energy Systems, e-Energy '19, {Association for Computing Machinery}, New York, NY, USA, 2019, pp. 1--11.
\newblock \href {https://doi.org/10.1145/3307772.3328314} {\path{doi:10.1145/3307772.3328314}}.

\bibitem{Ruan.2024}
J.~Ruan, Q.~Wang, S.~Chen, H.~Lyu, G.~Liang, J.~Zhao, Z.~Y. Dong, On vulnerability of renewable energy forecasting: Adversarial learning attacks, IEEE Transactions on Industrial Informatics 20~(3) (2024) 3650--3663.
\newblock \href {https://doi.org/10.1109/TII.2023.3313526} {\path{doi:10.1109/TII.2023.3313526}}.

\bibitem{Tang.2021}
N.~Tang, S.~Mao, R.~M. Nelms, Adversarial attacks to solar power forecast, in: 2021 IEEE Global Communications Conference (GLOBECOM), IEEE, 2021, pp. 1--6.
\newblock \href {https://doi.org/10.1109/GLOBECOM46510.2021.9685910} {\path{doi:10.1109/GLOBECOM46510.2021.9685910}}.

\bibitem{Heinrich.2024}
R.~Heinrich, C.~Scholz, S.~Vogt, M.~Lehna, Targeted adversarial attacks on wind power forecasts, Machine Learning 113~(2) (2024) 863--889.
\newblock \href {https://doi.org/10.1007/s10994-023-06396-9} {\path{doi:10.1007/s10994-023-06396-9}}.

\bibitem{Yang.2024}
L.~Yang, G.~Liang, Y.~Yang, J.~Ruan, P.~Yu, C.~Yang, Adversarial false data injection attacks on deep learning-based short-term wind speed forecasting, IET Renewable Power Generation 18~(7) (2024) 1370--1379.
\newblock \href {https://doi.org/10.1049/rpg2.12853} {\path{doi:10.1049/rpg2.12853}}.

\bibitem{Chen.2023}
Y.~Chen, M.~Sun, Z.~Chu, S.~Camal, G.~Kariniotakis, F.~Teng, Vulnerability and impact of machine learning-based inertia forecasting under cost-oriented data integrity attack, IEEE Transactions on Smart Grid 14~(3) (2023) 2275--2287.
\newblock \href {https://doi.org/10.1109/TSG.2022.3207517} {\path{doi:10.1109/TSG.2022.3207517}}.

\bibitem{Samad.2012}
T.~Samad, S.~Kiliccote, \href{https://www.sciencedirect.com/science/article/pii/S0098135412002347}{Smart grid technologies and applications for the industrial sector}, Computers {\&} Chemical Engineering 47 (2012) 76--84.
\newblock \href {https://doi.org/10.1016/j.compchemeng.2012.07.006} {\path{doi:10.1016/j.compchemeng.2012.07.006}}.
\newline\urlprefix\url{https://www.sciencedirect.com/science/article/pii/S0098135412002347}

\bibitem{Scholtz.2024}
E.~Scholtz, A.~Oudalov, I.~Harjunkoski, Power systems of the future, Computers {\&} Chemical Engineering 180 (2024) 108460.
\newblock \href {https://doi.org/10.1016/j.compchemeng.2023.108460} {\path{doi:10.1016/j.compchemeng.2023.108460}}.

\bibitem{Bree.2019}
L.~C. Br{\'e}e, K.~Perrey, A.~Bulan, A.~Mitsos, Demand side management and operational mode switching in chlorine production, AIChE Journal 65~(7) (2019).
\newblock \href {https://doi.org/10.1002/aic.16352} {\path{doi:10.1002/aic.16352}}.

\bibitem{Otashu.2019}
J.~I. Otashu, M.~Baldea, \href{https://www.sciencedirect.com/science/article/pii/S0098135418304411}{Demand response-oriented dynamic modeling and operational optimization of membrane-based chlor-alkali plants}, Computers {\&} Chemical Engineering 121 (2019) 396--408.
\newblock \href {https://doi.org/10.1016/j.compchemeng.2018.08.030} {\path{doi:10.1016/j.compchemeng.2018.08.030}}.
\newline\urlprefix\url{https://www.sciencedirect.com/science/article/pii/S0098135418304411}

\bibitem{Roben.2022}
F.~T.~C. R{\"o}ben, D.~Liu, M.~A. Reuter, M.~Dahmen, A.~Bardow, The demand response potential in copper production, Journal of cleaner production 362 (2022) 132221.

\bibitem{Schafer.2019}
P.~Sch{\"a}fer, H.~G. Westerholt, A.~M. Schweidtmann, S.~Ilieva, A.~Mitsos, Model-based bidding strategies on the primary balancing market for energy-intense processes, Computers {\&} Chemical Engineering 120 (2019) 4--14.
\newblock \href {https://doi.org/10.1016/j.compchemeng.2018.09.026} {\path{doi:10.1016/j.compchemeng.2018.09.026}}.

\bibitem{Zhang.2015}
Q.~Zhang, I.~E. Grossmann, C.~F. Heuberger, A.~Sundaramoorthy, J.~M. Pinto, Air separation with cryogenic energy storage: Optimal scheduling considering electric energy and reserve markets, AIChE Journal 61~(5) (2015) 1547--1558.
\newblock \href {https://doi.org/10.1002/aic.14730} {\path{doi:10.1002/aic.14730}}.

\bibitem{Caspari.2019}
A.~Caspari, C.~Offermanns, P.~Sch{\"a}fer, A.~Mhamdi, A.~Mitsos, A flexible air separation process: 2. optimal operation using economic model predictive control, AIChE Journal 65~(11) (2019) e16721.
\newblock \href {https://doi.org/10.1002/aic.16721} {\path{doi:10.1002/aic.16721}}.

\bibitem{HubertHadera.2015}
{Hubert Hadera}, {Iiro Harjunkoski}, {Guido Sand}, {Ignacio E. Grossmann}, {Sebastian Engell}, \href{https://www.sciencedirect.com/science/article/pii/S0098135415000472}{Optimization of steel production scheduling with complex time-sensitive electricity cost}, Computers {\&} Chemical Engineering 76 (2015) 117--136.
\newblock \href {https://doi.org/10.1016/j.compchemeng.2015.02.004} {\path{doi:10.1016/j.compchemeng.2015.02.004}}.
\newline\urlprefix\url{https://www.sciencedirect.com/science/article/pii/S0098135415000472}

\bibitem{Cramer.2024}
E.~Cramer, J.~Gao, A black-box adversarial attack on demand side management, Computers {\&} Chemical Engineering 186 (2024) 108681.
\newblock \href {https://doi.org/10.1016/j.compchemeng.2024.108681} {\path{doi:10.1016/j.compchemeng.2024.108681}}.

\bibitem{Ye.2025}
X.~Ye, W.~Tang, Process resilience under optimal data injection attacks, AIChE Journal (2025).
\newblock \href {https://doi.org/10.1002/aic.18896} {\path{doi:10.1002/aic.18896}}.

\bibitem{Parker.2023}
S.~Parker, Z.~Wu, P.~D. Christofides, Cybersecurity in process control, operations, and supply chain, Computers {\&} Chemical Engineering 171 (2023) 108169.
\newblock \href {https://doi.org/10.1016/j.compchemeng.2023.108169} {\path{doi:10.1016/j.compchemeng.2023.108169}}.

\bibitem{Zhang.2024}
G.~Zhang, B.~Sikdar, A novel adversarial fdi attack and defense mechanism for smart grid demand-response mechanisms, IEEE Transactions on Industrial Cyber-Physical Systems 2 (2024) 380--390.
\newblock \href {https://doi.org/10.1109/TICPS.2024.3448380} {\path{doi:10.1109/TICPS.2024.3448380}}.

\bibitem{Maciejowska.2026}
K.~Maciejowska, A.~Lipiecki, B.~Uniejewski, \href{https://www.sciencedirect.com/science/article/pii/S0196890426003778}{Statistical and economic evaluation of forecasts in electricity markets: Beyond rmse and mae}, Energy Conversion and Management 356 (2026) 121408.
\newblock \href {https://doi.org/10.1016/j.enconman.2026.121408} {\path{doi:10.1016/j.enconman.2026.121408}}.
\newline\urlprefix\url{https://www.sciencedirect.com/science/article/pii/S0196890426003778}

\bibitem{Zareipour.2009}
H.~Zareipour, C.~A. Canizares, K.~Bhattacharya, Economic impact of electricity market price forecasting errors: A demand-side analysis, IEEE Transactions on Power Systems 25~(1) (2009) 254--262.

\bibitem{Schafer.2020}
P.~Sch{\"a}fer, T.~M. Daun, A.~Mitsos, Do investments in flexibility enhance sustainability? a simulative study considering the german electricity sector, AIChE Journal 66~(11) (2020).
\newblock \href {https://doi.org/10.1002/aic.17010} {\path{doi:10.1002/aic.17010}}.

\bibitem{Germscheid.2022}
S.~H.~M. Germscheid, A.~Mitsos, M.~Dahmen, Demand response potential of industrial processes considering uncertain short--term electricity prices, AIChE Journal 68~(11) (2022).
\newblock \href {https://doi.org/10.1002/aic.17828} {\path{doi:10.1002/aic.17828}}.

\bibitem{Nowotarski.2016}
J.~Nowotarski, R.~Weron, \href{https://www.sciencedirect.com/science/article/pii/S014098831630127X}{On the importance of the long-term seasonal component in day-ahead electricity price forecasting}, Energy Economics 57 (2016) 228--235.
\newblock \href {https://doi.org/10.1016/j.eneco.2016.05.009} {\path{doi:10.1016/j.eneco.2016.05.009}}.
\newline\urlprefix\url{https://www.sciencedirect.com/science/article/pii/S014098831630127X}

\bibitem{Trebbien.2023}
J.~Trebbien, L.~{Rydin Gorj{\~a}o}, A.~Praktiknjo, B.~Sch{\"a}fer, D.~Witthaut, Understanding electricity prices beyond the merit order principle using explainable ai, Energy and AI 13 (2023) 100250.
\newblock \href {https://doi.org/10.1016/j.egyai.2023.100250} {\path{doi:10.1016/j.egyai.2023.100250}}.

\bibitem{Zhang.2020}
R.~Zhang, G.~Li, Z.~Ma, A deep learning based hybrid framework for day-ahead electricity price forecasting, IEEE Access 8 (2020) 143423--143436.
\newblock \href {https://doi.org/10.1109/ACCESS.2020.3014241} {\path{doi:10.1109/ACCESS.2020.3014241}}.

\bibitem{ENTSOEtransparencyplatform.2026}
{ENTSO-E transparency platform}, \href{https://transparency.entsoe.eu/}{Entso-e transparency platform} (2026).
\newline\urlprefix\url{https://transparency.entsoe.eu/}

\bibitem{Paszke.2019}
A.~Paszke, S.~Gross, F.~Massa, A.~Lerer, J.~Bradbury, G.~Chanan, T.~Killeen, Z.~Lin, N.~Gimelshein, L.~Antiga, et~al., Pytorch: An imperative style, high-performance deep learning library, Advances in neural information processing systems 32 (2019).

\bibitem{Pedregosa.2011}
F.~Pedregosa, G.~Varoquaux, A.~Gramfort, V.~Michel, B.~Thirion, O.~Grisel, M.~Blondel, P.~Prettenhofer, R.~Weiss, V.~Dubourg, J.~Vanderplas, A.~Passos, D.~Cournapeau, M.~Brucher, M.~Perrot, E.~Duchesnay, Scikit-learn: Machine learning in python, Journal of Machine Learning Research 12 (2011) 2825--2830.

\bibitem{Kurakin.2018}
A.~Kurakin, I.~Goodfellow, S.~Bengio, \href{http://arxiv.org/pdf/1607.02533}{Adversarial examples in the physical world}, in: R.~V. Yampolskiy (Ed.), Artificial Intelligence Safety and Security, Chapman {\&} Hall/CRC artificial intelligence and robotics series, {Chapman and Hall/CRC}, First edition. | Boca Raton, FL : CRC Press/Taylor {\&} Francis Group, 2018., 2018.
\newline\urlprefix\url{http://arxiv.org/pdf/1607.02533}

\bibitem{KavyaGupta.2021}
{Kavya Gupta}, {Beatrice Pesquet-Popescu}, {Fateh Kaakai}, {Jean-Christophe Pesquet}, {Fragkiskos D. Malliaros}, \href{https://centralesupelec.hal.science/hal-03527640/}{An adversarial attacker for neural networks in regression problems}, 2021.
\newline\urlprefix\url{https://centralesupelec.hal.science/hal-03527640/}

\bibitem{Biggio.2013}
B.~Biggio, I.~Corona, D.~Maiorca, B.~Nelson, N.~{\v{S}}rndi{\'c}, P.~Laskov, G.~Giacinto, F.~Roli, Evasion attacks against machine learning at test time, in: D.~Hutchison, T.~Kanade, J.~Kittler, J.~M. Kleinberg, F.~Mattern, J.~C. Mitchell, M.~Naor, O.~Nierstrasz, C.~{Pandu Rangan}, B.~Steffen, M.~Sudan, D.~Terzopoulos, D.~Tygar, M.~Y. Vardi, G.~Weikum, C.~Salinesi, M.~C. Norrie, {\'O}.~Pastor (Eds.), Advanced information systems engineering, Vol. 7908 of Lecture Notes in Computer Science, Springer, New York, 2013, pp. 387--402.
\newblock \href {https://doi.org/10.1007/978-3-642-40994-3{\textunderscore }25} {\path{doi:10.1007/978-3-642-40994-3{\textunderscore }25}}.

\bibitem{Barth.2018}
L.~Barth, N.~Ludwig, E.~Mengelkamp, P.~Staudt, A comprehensive modelling framework for demand side flexibility in smart grids, Computer Science-Research and Development 33~(1) (2018) 13--23.

\bibitem{Wanapinit.2021}
N.~Wanapinit, J.~Thomsen, C.~Kost, A.~Weidlich, An milp model for evaluating the optimal operation and flexibility potential of end-users, Applied Energy 282 (2021) 116183.

\bibitem{JoannahI.Otashu.2018}
{Joannah I. Otashu}, {Michael Baldea}, \href{https://www.sciencedirect.com/science/article/pii/S0306261918303647}{Grid-level ``battery'' operation of chemical processes and demand-side participation in short-term electricity markets}, Applied Energy 220 (2018) 562--575.
\newblock \href {https://doi.org/10.1016/j.apenergy.2018.03.034} {\path{doi:10.1016/j.apenergy.2018.03.034}}.
\newline\urlprefix\url{https://www.sciencedirect.com/science/article/pii/S0306261918303647}

\bibitem{Daryanian.2002}
B.~Daryanian, R.~E. Bohn, R.~D. Tabors, Optimal demand-side response to electricity spot prices for storage-type customers, IEEE Transactions on Power Systems 4~(3) (2002) 897--903.

\bibitem{Gellings.1985}
C.~W. Gellings, The concept of demand-side management for electric utilities, Proceedings of the IEEE 73~(10) (1985) 1468--1470.
\newblock \href {https://doi.org/10.1109/PROC.1985.13318} {\path{doi:10.1109/PROC.1985.13318}}.

\bibitem{Bynum.2021}
M.~L. Bynum, G.~A. Hackebeil, W.~E. Hart, C.~D. Laird, B.~L. Nicholson, J.~D. Siirola, J.-P. Watson, D.~L. Woodruff, Pyomo --- Optimization Modeling in Python, Vol.~67, {Springer International Publishing}, Cham, 2021.
\newblock \href {https://doi.org/10.1007/978-3-030-68928-5} {\path{doi:10.1007/978-3-030-68928-5}}.

\bibitem{GurobiOptimization.}
L.~L. {Gurobi Optimization}, \href{https://www.gurobi.com/}{Gurobi optimizer refereance manual}.
\newline\urlprefix\url{https://www.gurobi.com/}

\bibitem{Nocedal.2006}
J.~Nocedal, S.~J. Wright, Numerical optimization, Springer, 2006.

\bibitem{Mucci.2023}
S.~Mucci, A.~Mitsos, D.~Bongartz, \href{https://www.sciencedirect.com/science/article/pii/S2352152X2302011X}{Cost-optimal power-to-methanol: Flexible operation or intermediate storage?}, Journal of Energy Storage 72 (2023) 108614.
\newblock \href {https://doi.org/10.1016/j.est.2023.108614} {\path{doi:10.1016/j.est.2023.108614}}.
\newline\urlprefix\url{https://www.sciencedirect.com/science/article/pii/S2352152X2302011X}

\end{thebibliography}




\end{document}